\documentclass[sigconf]{acmart}
\renewcommand\footnotetextcopyrightpermission[1]{}
\usepackage{pifont}
\usepackage{booktabs}
\usepackage{makecell}
\usepackage{algorithm}
\usepackage{algpseudocode}
\usepackage{amsmath}
\usepackage{amsfonts}

\usepackage{graphicx}
\usepackage{booktabs}
\usepackage{multirow}

\usepackage{subcaption}
\usepackage{float}

\AtBeginDocument{%
  }

\setcopyright{acmlicensed}
\copyrightyear{2018}
\acmYear{2018}
\acmDOI{XXXXXXX.XXXXXXX}

\acmConference[Arxiv' 26]{Make sure to enter the correct
  conference title from your rights confirmation email}{2026}{Under Review}
\acmISBN{978-1-4503-XXXX-X/18/06}

\begin{document}

\title[]{CooperLLM: Cloud–Edge–End Cooperative Federated Fine-tuning for LLMs via ZOO-based Gradient Correction}

\author{He Sun}
\affiliation{%
  \institution{University of Science and Technology of China}
  \city{Hefei}
  \country{China}
}
\email{hesun@mail.ustc.edu.cn}

\author{Jinrui Zhou}
\affiliation{%
  \institution{University of Science and Technology of China}
  \city{Hefei}
  \country{China}
}
\email{zzkevin@mail.ustc.edu.cn}

\author{Li Li}
\affiliation{
  \institution{University of Macau}
  \city{Macau SAR}
  \country{China}
}
\email{llili@um.edu.mo}

\author{Mingjun Xiao}
\affiliation{
  \institution{University of Science and Technology of China}
  \city{Hefei}
  \country{China}
}
\email{xiaomj@ustc.edu.cn}










\begin{abstract}
Large Language Models (LLMs) have achieved remarkable success across numerous NLP tasks, yet fine-tuning them on resource-constrained mobile devices due to privacy and LLM personalization remains a major challenge due to their prohibitive memory and computational requirements. Federated Learning (FL) offers a privacy-preserving paradigm for distributed fine-tuning by keeping user data local. However, conventional FL approaches either rely on full backpropagation (BP), which incurs substantial memory of intermediate data and heavy GPU consumption, or adopt zeroth-order optimization (ZOO), which removes BP but suffers from high gradient variance and reduced accuracy due to random, non-directional perturbations.

To overcome these limitations, we propose CooperLLM, a cloud-assisted edge–end cooperative federated fine-tuning framework that combines the efficiency of ZOO with the accuracy of gradient rectification (ZGR). In CooperLLM, mobile clients perform lightweight ZOO-based updates on private datasets, while the cloud simultaneously conducts BP-based fine-tuning on auxiliary public datasets in the same downstream area. The cloud then extracts gradient subspaces and injects guided gradient knowledge back into the FL system to rectify the local zeroth-order updates. This cloud-guided gradient correction significantly improves convergence stability and accuracy while preserving user data privacy. To further address system bottlenecks that arise from frequent multi-tier communication between cloud and end devices and limited on-device memory, CooperLLM introduces two optimization controllers to orchestrate computation and communication across these heterogeneous devices. The System-level Pipeline Controller (SPC) overlaps cloud sampling, transmission, and client computation in a layer-wise pipeline, hiding communication latency and reducing memory usage. The Data Transmission Controller (DTC) adaptively compresses guided perturbations to ensure transmission time never exceeds computation time. Together, they enable latency-balanced and memory-efficient training across cloud, edge, and mobile tiers. We implement CooperLLM based on a representative federated NLP framework and evaluate it on multiple Transformer-based models and datasets. Experiments show that CooperLLM reduces on-device memory by up to 86.4\%, accelerates convergence by $8.8\times$, and improves accuracy by up to 10 percentage points over state-of-the-art ZOO-based baselines, enabling practical and privacy-preserving fine-tuning of LLMs on edge devices.
\end{abstract}




\maketitle

\section{Introduction}
In recent years, Large Language Models (LLMs) have experienced unprecedented development \cite{chang2024survey, yang2024harnessing} and have seen phenomenal mobile applications in various fields \cite{jin2023emsassist,lu2025mobiedit,yin2024elms}. Fine-tuning is an efficient technique for improving task adaptability and performance in various downstream natural language tasks using some domain-specific datasets \cite{zhang2023instruction,hu2021lora,chen2022revisiting,dettmers2024qlora}. However, these datasets are distributed across various mobile devices, which are private for the mobile users and cannot be leaked \cite{bai2024federated,villalobos2022will,khan2024float}.
Given data privacy concerns, Federated Learning (FL) has emerged as an ideal technique, enabling distributed cooperative training of models without transmitting local data \cite{mcmahan2017communication,kairouz2021advances,zhao2018federated} on devices.

While it offers many advantages, fine-tuning LLMs on resource-constrained mobile devices often faces severe memory bottlenecks, which in turn limit the participation rate of such devices \cite{wu2025survey,panchal2024thinking,xu2024fwdllm,cai2023federated,cai2023efficient}. For example, fine-tuning a LLaMA-7B model \cite{touvron2023llama} with a context length of 1K context and a batch size of 1 requires approximately 110 GB of memory. Furthermore, the memory footprint grows substantially as the context length increases. With the same batch size, an 8K context demands around 170 GB of memory. Nevertheless, mobile devices typically have only $4\sim16$ GB of memory \cite{scientiamobile2022ram}, which is far below the required capacity. Currently, some efforts leverage Parameter Efficient Fine-Tuning (PEFT) to address this issue by reducing the number of parameters updated \cite{hu2021lora,babakniya2023slora,zhang2023fedpetuning,pfeiffer2020adapterhub}. However, PEFT methods do not reduce the memory footprint of activations, which grows proportionally with the context length. For instance, with a batch size of 1, fine-tuning the Llama-7B model also requires 25.6 GB of memory for a 1K-token context and 87 GB for an 8K-token context, which still makes such fine-tuning infeasible on mobile devices (more details see Sec. \ref{Sec: MBD}). 


To find an efficient solution to reduce the memory footprint, an emerging method is proposed to replace the memory-intensive backpropagation (BP) required for first-order gradient computation methods with a BP-free optimizer, well known as Zeroth-Order Optimization (ZOO) \cite{fang2022communication}. Several existing works have leveraged BP-Free \cite{xu2024fwdllm,baydin2022gradients,ren2022scaling,liu2020primer}, particularly ZOO-based methods \cite{malladi2023fine,zhangrevisiting,qinfederated}, to improve memory utilization and communication efficiency. However, these methods suffer from two inherent limitations due to their gradient approximation mechanism \cite{tan2025harmony}. First, the zeroth-order gradient estimator exhibits high variance, typically $O(n)$ for $ n$-dimensional parameters \cite{nesterov2017random}, necessitating significantly more iterations (around $4.5\sim5.5\times$ in Fig. \ref{fig:compare_fo_zo}) than BP-based methods to converge, which is even worse for resource-constrained mobile devices. Second, the non-directional perturbation in ZOO introduces bias into gradient estimates, which is proportional to perturbation radius, causing cumulative deviations that degrade final model accuracy ($3\sim5\%$ accuracy reduction in Fig. \ref{fig:compare_fo_zo}) \cite{liu2018zeroth}.

To overcome these limitations, we reveal a key insight through theoretical analysis: introducing BP-based gradient rectification during each zeroth-order optimization round can significantly accelerate convergence and improve the model accuracy (See details in Sec. \ref{Sec:insight}). Based on this key insight, we can use the cloud-assistant pattern \cite{wang2024end} to complete this process. Specifically, we propose CooperLLM, an efficient, novel cloud-assisted edge-end cooperative federated fine-tuning framework for LLMs using zeroth-order optimization with gradient rectification. In CooperLLM, edge servers orchestrate the FL system with mobile devices, where these resource-constrained end devices perform ZOO-based on local privacy datasets for local fine-tuning; simultaneously, the cloud conducts BP-based fine-tuning on public auxiliary datasets; at synchronized intervals, the cloud infuses calibrated gradient knowledge into the FL system to rectify the ZOO process.


However, architecting this novel learning paradigm introduces inherent complexities, which give rise to two fundamental challenges. First, End devices have no visibility into cloud-side information, while client-side ZOO updates are high-variance; naively interpolating cloud BP directions with ZOO directions ignores parameter importance and yields suboptimal guidance. Thus, how to leverage limited BP gradients from the cloud to efficiently and accurately rectify ZOO gradients on clients remains a significant challenge. Second, Any BP-based gradients sent from the cloud, whether raw gradients or their surrogates, remains comparable in scale to full model parameters, incurring heavy cloud–edge–end communication, extra client memory pressure, and pipeline stalls. Therefore, optimizing communication latency while orchestrating computation and communication pipelines across heterogeneous devices becomes another key challenge.

To address the first challenge, we propose a cloud-assisted ZOO with the Gradient Rectification (ZGR) mechanism. During cloud-side BP-based fine-tuning, we maintain a gradient subspace and then sample guided perturbations from this subspace to rectify the forward-based zeroth-order perturbation sampling at end devices. In this way, end devices can sample gradient directions more accurately compared to conventional random perturbations. Furthermore, we adaptively tune the ZGR parameter based on the relative scales of BP-based and ZOO-based perturbations, enabling more stable and robust optimization during fine-tuning. 


To tackle the second challenge, we design two system-level optimization controllers: the System-level Pipeline Controller (SPC) and the Data Transmission Controller (DTC). In CooperLLM, the cloud has to transmit guided perturbations of a scale comparable to model weights to the FL system (both the edge server and end clients), introducing severe latency and memory bottlenecks. SPC addresses this issue by organizing the sampling, transmission, and application of guided perturbations in a layer-wise pipelined manner, which overlaps cloud sampling, transmission, and client computation across layers. This design significantly hides communication latency and reduces client memory usage, as only one layer of perturbations needs to be stored at a time. In addition, we replace transmitting gradients from clients to the edge server with sending only a constant-sized loss value. The edge server then combines guided perturbations with seed-generated local perturbations to compute the gradients. By offloading computation and storage of gradients from clients to the edge server (the overhead is negligible for the edge server, see details in \textit{Advanced Difference of Workflow} in Sec. \ref{Sec:overview}), this design reduces client memory usage and computational overhead while also lowering data transmission volume. Building on SPC, DTC further reduces transmission latency through quantization and compression. By adaptively adjusting the compression ratio based on observed computation delays, DTC ensures that communication can be fully hidden within computation, thereby minimizing end-to-end system latency while further lowering client memory footprint.

We implement CooperLLM based on a representative federated learning framework \cite{lin2021fednlp}, and evaluate it on multiple Transformer-based models across diverse datasets. Experimental results show that CooperLLM reduces memory consumption by up to 260\% compared to BP-based baselines, while achieving a 1.6× faster convergence speed than ZOO-based baselines, all without sacrificing model accuracy.

The main contributions are summarized as follows:

(1) We propose CooperLLM, the first efficient cloud-assisted edge-end cooperative federated fine-tuning framework with zeroth-order gradients rectification for large language models, which reduces memory footprint on end devices while accelerating convergence and the overall fine-tuning process. 

(2) We design a rectificatory gradient sampling mechanism that leverages cloud-side gradient subspaces to guide zeroth-order updates on end devices, improving gradient approximation and convergence efficiency.

(3) We introduce two system-level controllers to overlap computation with communication and adaptively compress data, effectively reducing latency in cloud–edge–end collaboration and memory footprint of end devices.

(4) We implement CooperLLM and evaluate it across multiple datasets and Transformer models, showing up to an 86.37\% reduction in end-device memory footprint compared to BP-based baselines, an $8.8\times$ faster convergence speed, and up to a 10-percentage-point improvement in accuracy on complex tasks over state-of-the-art ZOO-based methods.

\section{Background and Motivations}\label{Bg_Mo}
\subsection{Federated fine-tuning for LLM}\label{Sec: MBD}
Large Language Models have emerged as transformative breakthroughs, demonstrating exceptional performance across diverse domains. To adapt these foundation models to downstream tasks, fine-tuning pre-trained LLMs with domain-specific data has become prevalent. However, such data often contains sensitive and confidential information that cannot be exposed. Federated Learning for LLMs (FedLLM) thus presents a promising paradigm, enabling collaborative model training without sharing private local data \cite{cai2023efficient,cai2023federated,xu2024fwdllm,chen2023federated,zhang2024towards,wu2024fedbiot,kuang2024federatedscope}. 

The primary challenge in deploying FedLLM lies in bridging the substantial resource gap between massive model training demands and severely constrained mobile devices, well known as the "memory wall". While existing approaches attempt to mitigate this through Parameter-Efficient Fine-Tuning (PEFT) techniques (e.g., LoRA \cite{hu2021lora,chenlonglora,dettmers2024qlora,babakniya2023slora}, Adapters \cite{pfeiffer2020adapterhub}) or backpropagation-free (BP-free) optimization, they face fundamental limitations: PEFT methods fail to entirely overcome the memory wall during long-context activation, and BP-free alternatives incur significant performance degradation and training inefficiency. 


\subsection{Memory Breakdown Analysis of Fine-tuning}\label{Sec:Mem_Break}
In this subsection, we investigate how the memory wall impacts FedLLM under varying context lengths. To empirically explore this question, we emulate a FedLLM environment configured with 200 clients featuring heterogeneous memory budgets: 10\% with 64GB, 15\% with 32GB, 20\% with 16GB, 25\% with 8GB, and 30\% with 4GB, inspired by NVIDIA Jetson series devices. We conduct experiments using the LLaMA-7B \cite{touvron2023llama} model on the LongAlign \cite{bai2024longalign} dataset. We use the Dirichlet distribution, as described in \cite{xu2022fedcorr}, and divide different context lengths to sample the data for each client.

\begin{figure}[tbp]
  \centering
  \includegraphics[width=\linewidth]{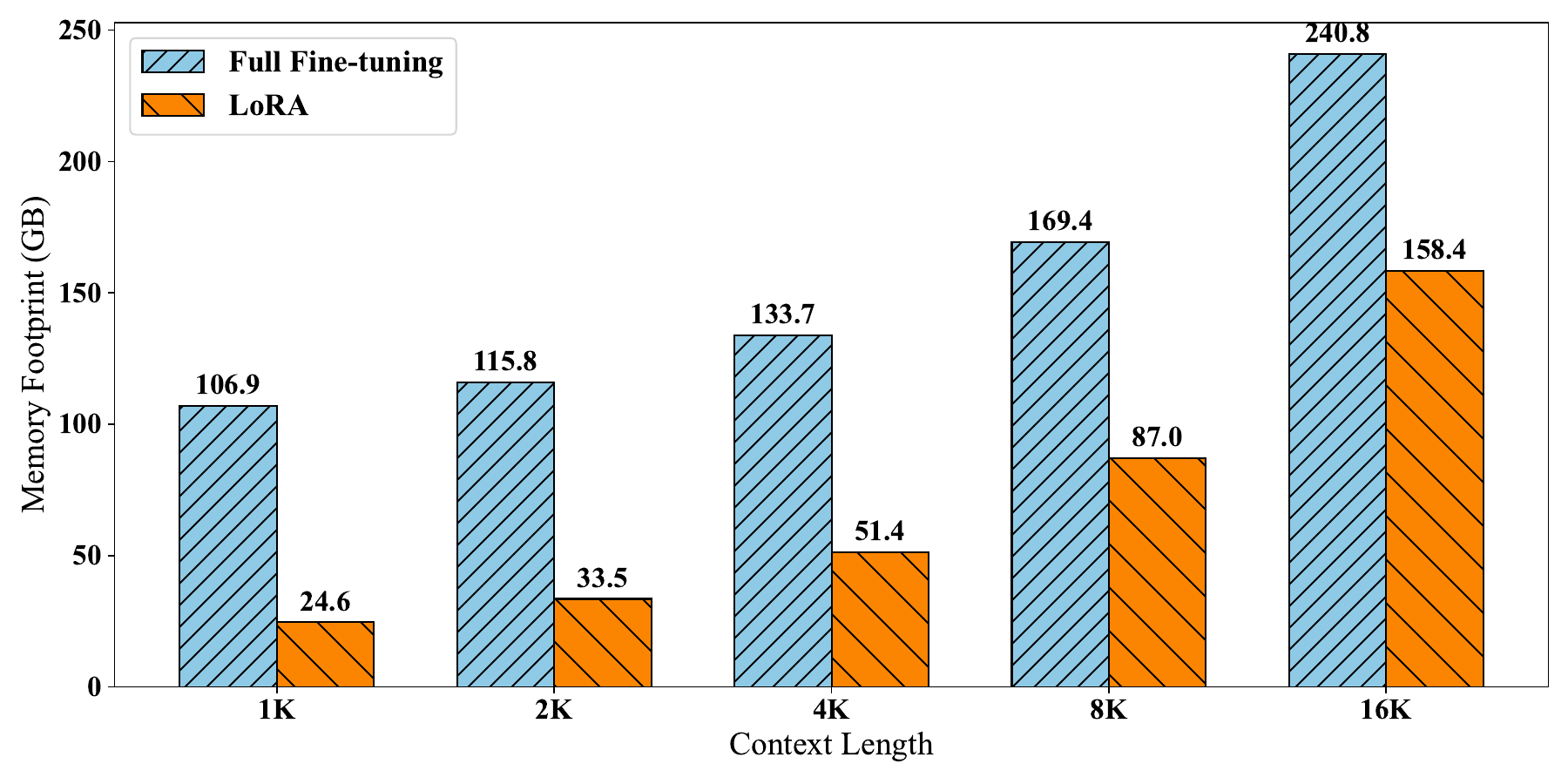}
  \caption{Memory breakdown between full fine-tuning and LoRA under varying prompt lengths with LlaMA-2-7B.}
  \label{fig:memory_bar}
\end{figure}

Figure \ref{fig:memory_bar} illustrates the memory footprint of full fine-tuning versus LoRA methods across different context lengths. Our analysis reveals that memory usage increases substantially with longer contexts in both methods, primarily because activation memory scales proportionally with context length. When the context length is merely 1K, full parameter fine-tuning of the Llama-7B model requires over 106GB of memory, far exceeding even high-end 64GB mobile devices. Under the same context length, the LoRA approach still demands approximately 24.6GB, surpassing the memory capacity of most mobile devices. Table \ref{tab:Par_rate} presents client participation rates in FedLLM under this configuration: full fine-tuning exhibits 0\% participation across all devices due to memory constraints, while LoRA achieves merely 25\% participation at 1K context and just 10\% at 4K context.

\begin{table}[tbp]
  \centering
  \caption{Partition Rate under Different Context Lengths using Llama-2-7B model and LongAlign dataset.}
  \label{tab:partition}
  \begin{tabular}{lccccc}
    \toprule
    Context Length & 1K & 2K & 4K & 8K & 16K \\
    \midrule
    Partition Rate with FT   & 0\% & 0\% & 0\% & 0\% & 0\% \\
    Partition Rate with LoRA & 25\% & 10\% & 10\% & 0\% & 0\% \\
    \bottomrule
  \end{tabular}
  \label{tab:Par_rate}
\end{table}

\subsection{Demystifying Existing Techniques of FedLLM}
In this subsection, we evaluate existing FedLLM techniques, analyzing their effectiveness in mitigating the memory wall and identifying their inherent limitations. Before commencing our analysis, we first dissect the memory footprint during LLM fine-tuning, which comprises four components: model parameters (stored in FP16, consuming $2P$ bytes), optimizer states (e.g., momentum and variance in AdamW, stored in FP32 and requiring $8P$ bytes), gradients (stored in FP32, occupying $4P$ bytes), and activations (stored in FP16, with total memory given by ($A = 2L[(V_{fwd}+V_{bwd}]L_{ctx}BD+5HL_{ctx}B$).  Here, $P$ denotes the number of trainable parameters, $L$ denotes the number of model layers, $L_{ctx}$ denotes the context length in tokens, $B$ denotes the batch size, $D$ denotes the hidden dimension size, $H$ denotes the number of attention heads, and $V_{fwd}, V_{bwd}$ denotes the number of variables stored during forward/backward propagation per layer. 
\begin{figure}[tbp]
  \centering
  \includegraphics[width=\linewidth]{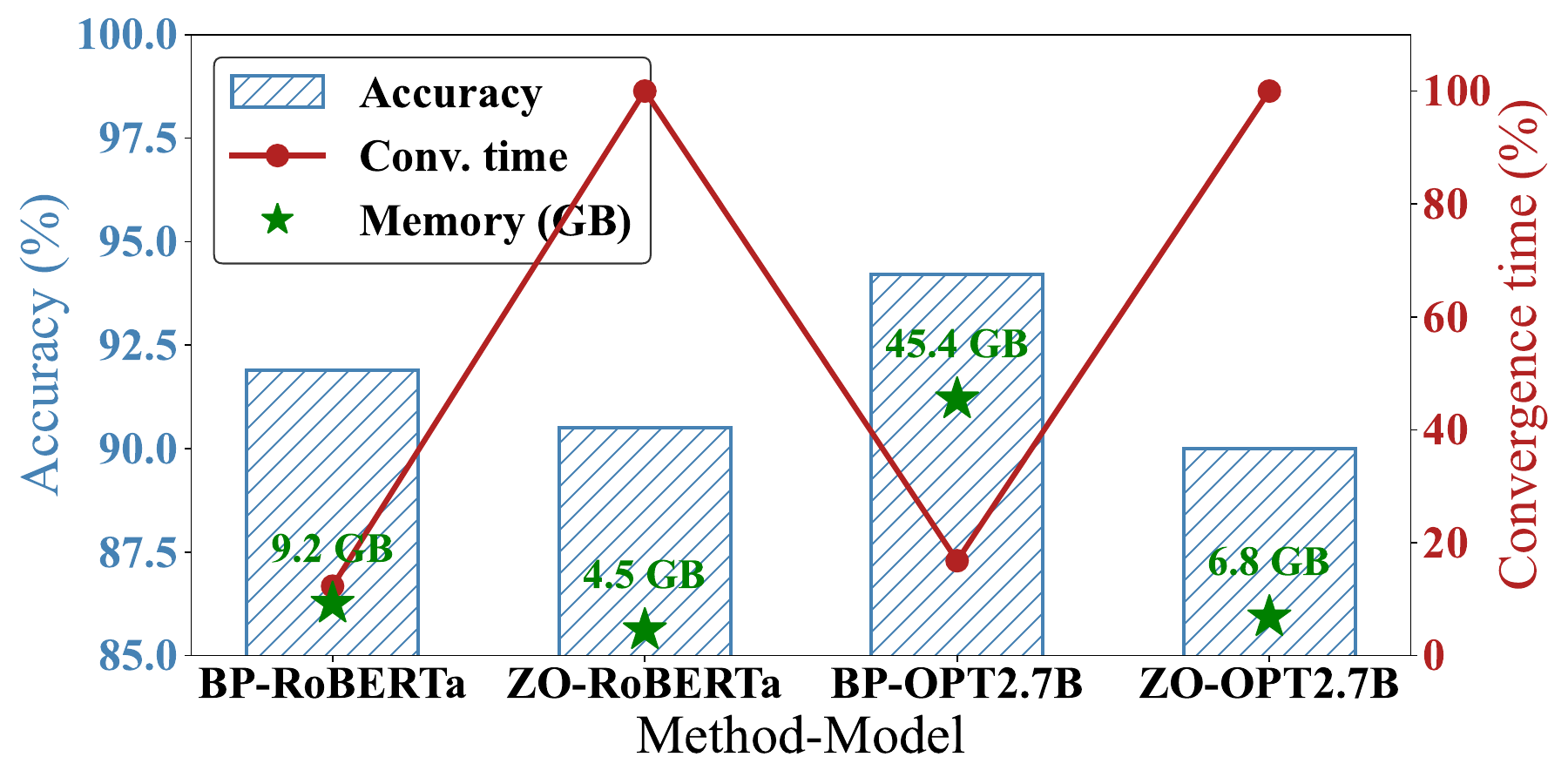}
  \caption{The Comparison of BP-based Method and ZOO-based Method on Model Performance and Convergence Time with different Models on SST-2 \cite{stanfordnlp-sst2} Datasets.}
  \label{fig:compare_fo_zo}
\end{figure}

\noindent\textbf{Parameter-efficient Fine-tuning.} 
Parameter-efficient Fine-tuning for LLMs aims to reduce memory consumption during training by minimizing the number of updated parameters, as exemplified by techniques like LoRA \cite{hu2021lora,dettmers2024qlora} and Adapters \cite{pfeiffer2020adapterhub,cai2023efficient}. LoRA-style methods introduce trainable low-rank matrices into both the attention and feed-forward layers, thereby significantly reducing the number of parameters that need to be updated during training. Similarly, Adapters-style methods primarily integrate compact, task-oriented modules into transformer layers, enabling efficient fine-tuning by freezing the main model and updating only the adapter parameters. However, the PEFT methods primarily reduce the memory footprint of optimizer states and gradients, bringing it down to roughly $1~2\%$ of the full parameter size (about 8P). However, they do not optimize the memory consumed by activations, which grows proportionally with the context length \cite{wang2025jenga}. 

Figure \ref{fig:memory_bar} shows that LoRA-style methods reduce memory usage by $34\%\sim77\%$ compared with full-parameter fine-tuning across context lengths from 1K to 16K. However, Table \ref{tab:Par_rate} shows the client participation rate when fine-tuning LLaMA-7B with LoRA-style methods in a federated learning setting. We observe that when the context length is below 4K, only $10\sim25\%$ of clients can participate, reflecting a low participation level. Once the context length exceeds 4K, the participation rate drops to nearly zero. As a result, PEFT methods still encounter severe memory bottlenecks when fine-tuning LLMs on mobile devices.

\noindent\textbf{Zeroth-order Optimization for Fine-tuning.}
Unlike BP-based PEFT methods, BP-free Zeroth-order Optimization (ZOO)-style methods \cite{fang2022communication,xu2024fwdllm,ling2024convergence,qinfederated} estimate gradient directions through random sampling and update parameters using only forward passes instead of BP. By eliminating the need to store optimizer states and activations from BP, these methods substantially reduce the memory footprint during LLM fine-tuning. Despite these advantages, ZOO-style methods require multiple random samples to decrease high variance when estimating gradient directions \cite{nesterov2017random}, and the non-directional perturbation in ZOO introduces bias into gradient estimates \cite{liu2018zeroth}. As a result, they converge more slowly and yield inferior model performance than BP-based approaches \cite{tan2025harmony}.

To empirically validate these viewpoints, we follow the experimental setting in \cite{tan2025harmony,zhangrevisiting} and compare BP-based and ZOO-based fine-tuning methods on the SST-2 \cite{stanfordnlp-sst2} dataset using RoBERTa \cite{liu2019roberta} and OPT-2.7B \cite{zhang2022opt} models. As shown in Fig.~\ref{fig:compare_fo_zo}, we evaluate their differences in terms of accuracy and convergence time. We find that, compared to BP-based methods, ZOO-based fine-tuning results in accuracy drops of approximately 1.6\% on RoBERTa and 4\% on OPT-2.7B, while convergence times increase by 88\% and 83\%, respectively. Additionally, the classical BP-free FwdLLM \cite{xu2024fwdllm} for FedLLM can cause approximate 5.8\% performance drop compared with the BP-based method as shown in \cite{zhan2025assyllm}. Although ZOO-style methods significantly reduce memory footprint (as shown in Fig. \ref{fig:compare_fo_zo}), the experiments confirm that they suffer from performance degradation and slower convergence.

\noindent\textbf{Summary.} Although these methods reduce memory consumption during fine-tuning to varying degrees, they either fail to address the memory bottleneck on mobile devices or incur accuracy loss and prolonged convergence. Moreover, system-level optimizations are still necessary to further accelerate the overall fine-tuning process. To overcome these limitations, we present a critical insight that motivates the design of our framework.

\subsection{The Insight and Challenges for CooperLLM}\label{Sec:insight}
To achieve higher participation rates in federated fine-tuning, we prioritize ZOO-style methods to reduce memory consumption. However, to address its limitation of accuracy degradation and slow convergence, we observe and argue an insight: if we introduce BP-based directional guidance and rectification into the random gradient sampling of ZOO can significantly improve gradient estimation accuracy and accelerate convergence. To validate this critical insight, we provide the evidence through theoretical analysis. 

\noindent \textbf{The theoretical feasibility analysis of gradient rectification.} We minimize a loss $f(\mathbf{x})$ that is $L$-smooth and satisfies the $\mu$-PL inequality \cite{karimi2016linear}. Let the ZOO two-point estimator be $\mathbf{g}_\text{Z}=\nabla f(\mathbf{x})+\boldsymbol{\zeta}$ with $\boldsymbol{\zeta}\sim N(0,\sigma^2)$, and suppose the BP gradient be $\mathbf{g}_{B}=\nabla f(\mathbf{x})+\mathbf{b}$ with $\mathbb{E}[\mathbf{b}]=\bar{\mathbf{b}}$, $\text{Var}(\mathbf{b})=\tau^2$.

The most innocent and straightforward idea is to take a weighted combination of $\mathbf{g}_{B}$ and $\mathbf{g}_{Z}$ as the updated rectificatory gradient. Every $T$ rounds we form the gradient rectification $\mathbf{g}_\text{R}=\lambda\mathbf{g}_{B}+(1-\lambda)\mathbf{g}_{Z}$, $0\le\lambda\le1$. The bias and variance of $\mathbf{g}_{R}$ is $\lambda\bar{\mathbf{b}}$ and $\text{Var}=\lambda^2\tau^2+(1-\lambda)^2d\sigma^2$, giving the $\text{MSE}(\lambda)=\lambda^2\bigl(\|\bar{\mathbf{b}}\|^2+\tau^2\bigr)+(1-\lambda)^2d\sigma^2$. Minimizing $\text{MSE}(\lambda)$ yields the optimal mixing weight $\lambda^*=\frac{d\sigma^2} {d\sigma^2+\|\bar{\mathbf{b}}\|^2+\tau^2}$. As for $\mu$-PL inequality, with diminishing step size $\eta_t=1/(\mu t)$, the expected sub-optimality satisfies
\begin{equation}\label{Eq:MSE}
\mathbb{E}[f(\mathbf{x}_t)-f^*]\le \mathcal{O}\!\left(\frac{L_0}{\mu^2}\frac{\text{MSE}(\lambda)}{t}\right).
\end{equation}

\noindent\textit{Convergence Rate}: Let the iteration complexities required $T_{Z}$ and $T_{R}$ for pure ZOO and our idea with an $\varepsilon$-accurate solution be:
\begin{equation}
T_{Z} = \frac{L_0}{\mu^2}\frac{d\sigma^2}{\varepsilon},
\qquad
T_{R} = \frac{L_0}{\mu^2}\frac{\operatorname{MSE}(\lambda^*)}{\varepsilon},
\end{equation}
where $L_0$ is the Lipschitz constant. Substituting the optimal weight $\lambda^*$ gives $\operatorname{MSE}(\lambda^*) = \frac{d\sigma^2\bigl(\|\bar{\mathbf{b}}\|^2+\tau^2\bigr)}{d\sigma^2+\|\bar{\mathbf{b}}\|^2+\tau^2}$. So the speed-up ratio becomes:
\begin{equation}
\frac{T_{Z}}{T_{R}}
= \frac{d\sigma^2}{\operatorname{MSE}(\lambda^*)}
= 1 + \frac{d\sigma^2}{\|\bar{\mathbf{b}}\|^2 + \tau^2}.    
\end{equation}
For any non-zero BP bias or variance ($\|\bar{\mathbf{b}}\|^2+\tau^2>0$), gradient rectification strictly accelerates convergence.  In the practical regime where $\|\bar{\mathbf{b}}\|^2+\tau^2 \ll d\sigma^2$, the speed-up ratio $\gg1$. Therefore, the gradient rectification significantly reduces the number of convergence rounds compared to pure ZOO.

\noindent\textit{Accuracy ratio}: Under a fixed $t$ rounds, according to Eq. (\ref{Eq:MSE}) and substituting the respective MSE terms, we can get: 
\begin{align}
\varepsilon_{Z} &= \frac{L\,d\sigma^2}{\mu^2 K}, \\[2pt]
\varepsilon_{R}  &= \frac{L\,d\sigma^2}{\mu^2 K}\left(1+\frac{\|\bar{\mathbf{b}}\|^2+\tau^2}{d\sigma^2}\right)^{-1}.
\end{align}
Hence, we can get the accuracy improvement ratio:
\begin{equation}
    \frac{\varepsilon_{R}}{\varepsilon_{Z}}
=\Bigl(1+\tfrac{\|\bar{\mathbf{b}}\|^2+\tau^2}{d\sigma^2}\Bigr)^{-1}<1,
\end{equation}
which shows that gradient rectification strictly outperforms pure ZOO in final accuracy for any non-zero BP bias or variance.

\noindent \textbf{Challenges of designing CooperLLM.} Building on the above insight, we run in the cloud a BP-based replica of the same model used in the FL system. At a fixed number of rounds, the cloud rectifies the ZOO-estimated gradient directions on end devices, ensuring more accurate gradient approximation and thereby improving both model accuracy and convergence speed. Motivated by this, we propose CooperLLM, an efficient cloud-assisted edge–end cooperative federated fine-tuning framework with zeroth-order gradient rectification for LLMs. However, designing CooperLLM introduces two challenges at both the algorithmic and system levels. \textbf{Challenge 1}: In the innocent idea discussed within the insight, end clients lack awareness of cloud-side information, making it difficult to effectively trade off between the BP-based gradient directions computed in the cloud and the ZOO-based gradients in the federated system, which also suffer from high variance. Moreover, simply weighting the two gradient directions to obtain a single directional vector fails to capture parameter importance, thereby limiting the ability to leverage the most informative directions for effective rectification. Therefore, the algorithmic challenge lies in how to leverage BP-based gradients to efficiently and accurately calibrate ZOO-based gradients. \textbf{Challenge 2}: In the CooperLLM framework, regardless of the form in which BP-based gradients in the cloud are transmitted, their size remains comparable to the full model parameters. This not only introduces significant communication overhead between the cloud and the FL system but also imposes additional memory usage pressure on resource-constrained client devices. Moreover, in this cloud-assisted FL framework, the interplay between device-side computations and inter-device data transmissions inevitably introduces additional latencies into the training process. Thus, the system-level challenge lies in how to effectively optimize and pipeline these operations across heterogeneous devices to accelerate the overall fine-tuning process.

\section{CooperLLM Design}
\subsection{Overview}\label{Sec:overview}
CooperLLM is a cloud-assisted federated fine-tuning framework that enables resource-constrained mobile devices to fine-tune large language models efficiently. The core idea is to leverage a cloud-side BP-based model to rectify the gradients of ZOO-based models within the FL system, thereby mitigating the low accuracy and slow convergence inherent to ZOO. Meanwhile, CooperLLM introduces system-level optimizations that pipeline computation and communication, effectively reducing latency and accelerating the overall federated fine-tuning process.
\begin{figure}[tbp]
  \centering
  \includegraphics[width=\linewidth]{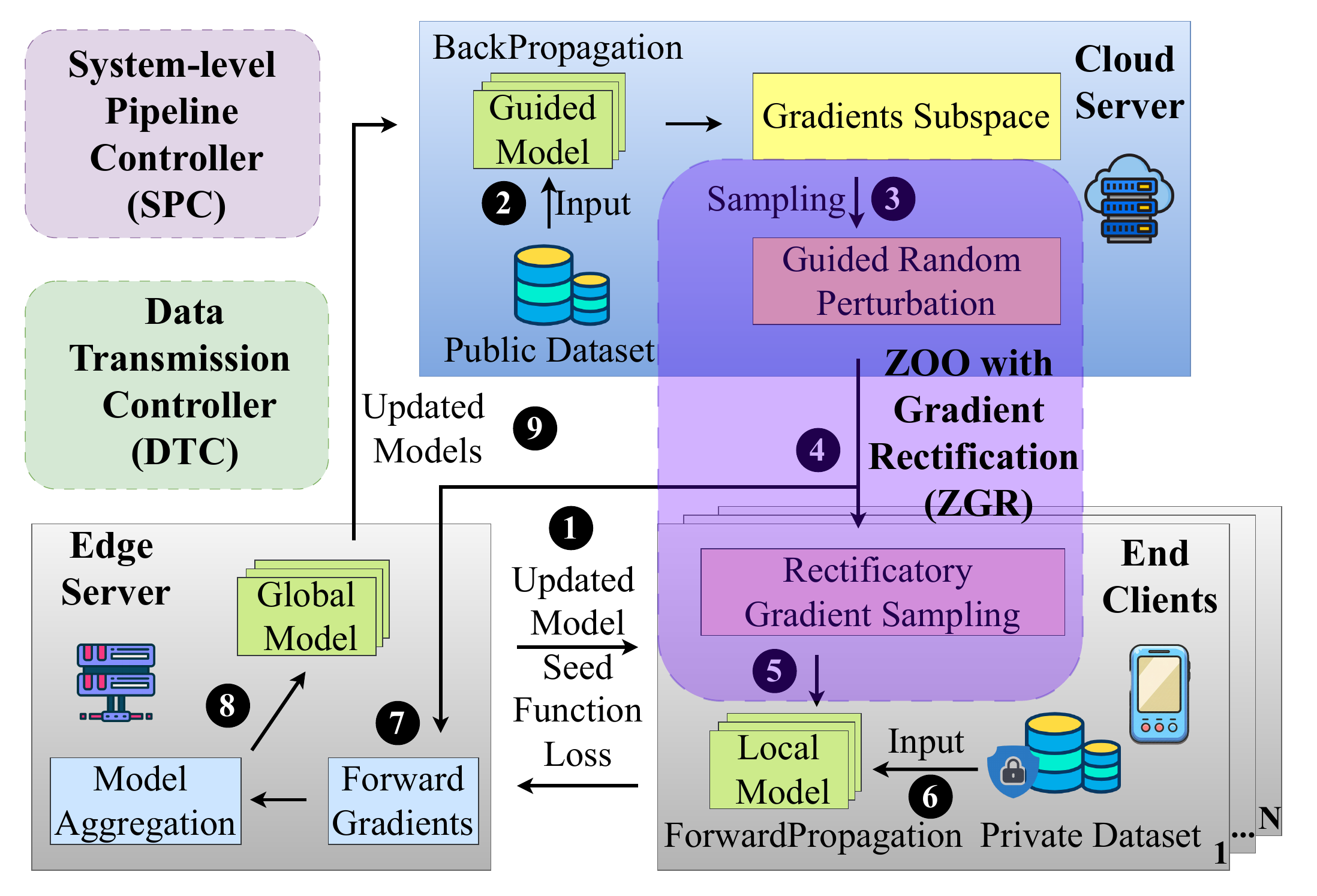}
  \caption{The Framework and Workflow of CooperLLM.}
  \label{fig:framwork}
\end{figure}

\noindent\textbf{CooperLLM Framework.} As shown in Figure \ref{fig:framwork}. CooperLLM is composed of three key components: the algorithmic design, Zeroth-order Optimization with Gradient Rectification (ZGR), and two system-level designs, the System-level Pipeline Controller (SPC) and the Data Transmission Controller (DTC). The core idea of ZGR is to leverage historical BP-trained gradients on the cloud to generate guided perturbations that reflect parameter importance, which are then used to rectify the ZOO gradients on end clients (a.k.a., clients). Instead of simply weighting BP-based and ZOO-based gradients, this approach enables more effective rectification by focusing on the most critical gradient parameters (Sec. \ref{Sec:ZGR}). The core idea of SPC is to handle the transmission of guided perturbations and their rectification layer by layer across the model, while executing transmission and computation in parallel. This reduces the memory footprint on clients and effectively hides communication latency (Sec. \ref{Sec:SPC}). The core idea of DTC is to further reduce transmission latency during parallel execution by compressing the transmitted data. DTC dynamically controls the compression ratio to ensure that transmission latency is effectively hidden within computation latency \ref{Sec:DTC}).

\noindent\textbf{Workflow.} Figure \ref{fig:framwork} illustrates the workflow of CooperLLM. The FL system is composed of edge servers and clients, where clients perform local updates on private data using ZOO-based training. Meanwhile, the cloud leverages BP-based training on public datasets from the same domain with the same model architecture. The detailed workflow is as follows: \ding{182} In each global round, the aggregator at the edge server distributes the latest updated model parameters and seeds to all available clients. \ding{183} Meanwhile, the cloud performs BP-based training on a public dataset while constructing a gradient subspace during the training process. \ding{184} It then samples guided perturbation with the same size as the updated LLM model parameter from this subspace and transmits it to each available client and the edge server at a regular interval. \ding{185} Each available client receives and stores the guided perturbation transmitted from the cloud, and keeps them locally in preparation for subsequent computations. \ding{186} Each available client independently generates local random perturbations (a.k.a. local perturbations) using the random seed. These locally generated perturbations are then computed with the guided perturbations received from the cloud to get the hybrid random perturbation (a.k.a., hybrid perturbations). The client directly applies the hybrid perturbations to its local model, thereby constructing the perturbed LLM. \ding{187} Each client performs forward passes on both the original model and the perturbed model using its local private dataset, and then derives the function loss from the difference between the two outputs. \ding{188} Each available client sends its locally function loss to the edge server. The edge server then combines this loss with the guided random perturbations (a.k.a., guided perturbations) to compute the forward gradients for each client. \ding{189} The edge server aggregates these forward gradients and applies the result to update the global model. \ding{190} Finally, the edge server uploads the updated model to the cloud for the next round of BP-based training. These steps are repeated until the model converges.

\noindent\textit{Advanced Difference of workflow.} Unlike conventional FL systems, CooperLLM does not compute forward gradients on the clients after local forward passes, as doing so would impose additional overhead on memory-constrained devices. Instead, each client transmits only a constant-sized function loss to the edge server rather than the gradients themselves. The edge server then derives the gradients accordingly. This design not only reduces the computation and memory footprint on clients but also the communication data volume, thereby mitigating latency. Note that the edge server only needs to maintain the guided perturbation and a single local perturbation to compute the forward gradient for each client sequentially. This introduces only one additional perturbation in terms of storage, which is well within the capacity of the edge server.

\subsection{Zeroth-order Optimization with Gradient Rectification}\label{Sec:ZGR}
\noindent\textbf{ZOO on Client.} Consider an FL system with $N$ clients $\{1,2,...,N\}$, where each client fine-tunes a pre-trained model $M^0 \in \mathbb{R}^d$ on its private dataset $D_i$ to collaboratively update a global model $W \in \mathbb{R}^d$. The objective is to minimize the total loss across all clients through federated fine-tuning:
\begin{equation} 
    \min_{W \in \mathbb{R}^d} f(W)|_{W^0}= \sum_{i=1}^{N} w_i \mathbb{E}_{x_i\sim D_i} L(W;x_i),
\end{equation}
where $L(W;x_i)$ is the loss of model $W$ on data $x_i$ sampled from $D_i$. $w_i$ is the aggregate weight of client $i$ and $\sum_{i=1}^N w_i = 1$. In BP-based approaches, such as Vanilla FL \cite{mcmahan2017communication}, each client $i$ performs several steps of stochastic gradient descent to update its local model: 
\begin{equation}
    W^r_{i,t+1} = W^r_{i,t} - \eta \cdot g^r_{i,t},
\end{equation}
where $r,t$ are the round and step numbers, $\eta, g^r_{i,t}$ are the learning rate and the gradient computed by $g^r_{i,t} = \nabla L(W^r_{i,t};x_i)$. After local training, each client $i$ sends its model $W^r_{i,t}$ to the server for aggregation. Unlike Vanilla FL \cite{mcmahan2017communication}, ZOO-based methods \cite{malladi2023fine,fang2022communication} estimate gradients by performing forward passes. The estimated gradient $\hat{g}^r_{i,t}$ is then computed from the difference in losses between these passes, combined with a randomly sampled perturbation. Here, we use antithetic sampling \cite{shapiro2003monte}: 
\begin{align}\label{Eq:ZOO}
    \hat{g}^r_{i,t} &= \frac{L(W^r_{i,t}+\epsilon z;x_i)-L(W^r_{i,t}-\epsilon z;x_i)}{2\epsilon} \cdot z 
\end{align}
where $z \in \mathbb{R}^d$, $z \sim N(0, I_d)$ and $\epsilon$ is the scale of perturbation. 

\noindent\textbf{Guided Perturbation Sampling on Cloud.} We conduct the BP-based training \cite{mcmahan2017communication} based on the same pre-trained model $W^0$ of LLM initially using the public dataset $D_p$. Every $\gamma$ rounds, the edge server transmits the updated model to the cloud for model update. At the end of each training round, the cloud stores the computed gradients to construct a subspace of gradients $\mathbb{G}_m$ \cite{maheswaranathan2019guided}. We denote the orthonormal basis of this subspace as $V\in\mathbb{R}^{m\times n}$. The guided perturbations are then sampled as $z_g\sim N(0,I_m)$, where $I_m = VV^T$. Then, the result of $Vz_g$ is transmitted to each available client and edge server.
\begin{algorithm}[t]
\caption{ZGR Algorithm}
\label{alg:zgr}
\begin{algorithmic}[1]
\Require Pre-trained model $W^0$; client datasets $\{D_i\}_{i=1}^N$;
         public dataset $D_p$; weights $\{w_i\}_{i=1}^N$; $\eta$; $\epsilon,\sigma$;
         guidance interval $\gamma$; subspace dim $m$; mixing ratio $\alpha$;
         shared random seed $s$.

\Statex \rule{\linewidth}{0.4pt}
\State \textbf{Cloud:}
\State Initialize global model $W^0$, set round $r\gets 0$.
    \State Perform BP on $D_p$ with $W^r$.
    \State Collect gradients to build basis $V_r\in\mathbb{R}^{d\times m}$.
\If{$r \bmod \gamma = 0$}                   \Comment{Guided round}
    \State Sample $z_g\sim\mathcal{N}(0,I_m)$;
           Broadcast guided perturbation $V_r z_g$ to all nodes (Clients and Edge Server).
\EndIf
\Statex \rule{\linewidth}{0.4pt}

\State \textbf{Clients (parallel):}
\State Receive $W^r$ from edge server.
\If{$r \bmod \gamma = 0$}
    \State Form hybrid perturbation
           $z_h\gets\sqrt{\tfrac{\alpha}{n}}z
                   +\sqrt{\tfrac{1-\alpha}{m}}V_r z_g$
           (see Eq. (\ref{Eq:perturbation})).
\Else
    \State Use local perturbation $z_h\gets z\sim\mathcal{N}(0,I_m)$.
\EndIf
\State Compute loss difference
       $\Delta_i\gets L(W^r+\epsilon z_h;x_i)-L(W^r-\epsilon z_h;x_i)$
       (See \ Eq. (\ref{Eq:ZOO})).
\State Send scalar $\Delta_i$ to edge server.
\Statex \rule{\linewidth}{0.4pt}

\State \textbf{Edge Server:}
\For{each client $i$}
    \State Regenerate identical $z_h$ via seed $s$.
    \State Recover gradient
           $\hat{g}_{i,t}^r\gets \frac{\Delta_i}{2\epsilon} \cdot z_h$ (See \ Eq. (\ref{Eq:grad})).
\EndFor
\State Aggregate
       $\hat{g}^r\gets\sum_{i=1}^{N} \sum_t w_i\hat{g}_{i,t}^r$.
\State Update $W^{r+1}\gets W^r - \eta\hat{g}^r$.
\State Broadcast $W^{r+1}$ to all clients.
\State $r\gets r+1$.
\end{algorithmic}
\end{algorithm}

\noindent\textbf{Cloud-assisted ZGR on Client.} 
First, during guided rounds $b\gamma,\ b\in Z^+$, each available client receives guided perturbations from the cloud and combines them with locally sampled perturbations to generate a hybrid perturbation $z_h$:
\begin{equation}\label{Eq:perturbation}
    z_h =  \sqrt{\frac{\alpha}{n}}z+\sqrt{\frac{1-\alpha}{m}}Vz_g,
\end{equation}
where $\alpha$ is the ZGR parameter. Next, each client performs a forward pass on its local private dataset $D_i$, using the perturbed local model $M^{b\gamma}_{i}$, to obtain the function value $f(M^{b\gamma}_{i})$. According to Eqs.(\ref{Eq:ZOO}), we can get:
\begin{align}\label{Eq:grad}
  \hat{g}^r_{i,t} &= \frac{L(W^r_{i,t}+ \epsilon z_h;x_i)-L(W^r_{i,t}- \epsilon z_h;x_i)}{2\epsilon} \cdot z_h
\end{align}
At this stage, as highlighted in the Advanced Difference of Workflow (Sec. \ref{Sec:overview}, a.k.a., ADW), each client transmits only the function loss, i.e., $[L(W^r_{i,t}+ \epsilon z_h;x_i)-L(W^r_{i,t}- \epsilon z_h;x_i)]$, to the edge server. In non-guided rounds, clients follow the same procedure but use locally sampled perturbations $z$. 

\noindent\textbf{Gradient Generation and Aggregation on Edge Server.} 
First, the edge server computes the gradient $\hat{g}^r_{i,t}$ for each client by combining its function loss with the corresponding perturbation according to Eqs. (\ref{Eq:grad}). Specifically, in non-guided rounds, the server regenerates the same local perturbation $z$ using the shared random seed to ensure consistency with the client. In guided rounds, however, it employs the hybrid perturbation $z_h$, which integrates the local perturbation $z$ with the guided perturbation received $z_g$ from the cloud using Eqs. (\ref{Eq:perturbation}). 
The edge server then computes the forward gradients using the hybrid perturbations and the function losses received from clients according to Eqs. (\ref{Eq:grad}). It subsequently aggregates the forward gradients and applies them to the global model $M$. During the aggregation process, we also adopt variance control techniques similar to those proposed in \cite{xu2024fwdllm, liu2018zeroth, maheswaranathan2019guided}. Finally, the latest updated model is transmitted to the cloud.

\noindent\textbf{Algorithm.} The whole process details are displayed in Algorithm \ref{alg:zgr}. which is divided into three parts (Cloud, Clients, Edge Server).










\subsection{System-level Pipeline Controller}\label{Sec:SPC}
In the ZGR module, during guided rounds, the cloud needs to transmit guided perturbations with a comparable size to the model weight to both the clients and the edge server. While the cloud-to-edge transmission latency can be effectively hidden by overlapping with client-side training and communication, the cloud-to-client transmission delay emerges as a critical bottleneck, significantly slowing down the overall training process. Moreover, clients must store the guided perturbations in order to construct the hybrid perturbation $g_h$, which imposes additional memory pressure on resource-constrained clients. To address this issue, we introduce the System-level Pipeline Controller (SPC).

\noindent\textbf{Layer-wise Latency Reduction.} SPC organizes the computation and transmission of guided perturbations layer by layer according to the model hierarchy. While sampling the guided perturbation for the second layer, the first layer's transmission of the guided perturbation runs in parallel. On the client side, when the first-layer hybrid perturbation is computed and applied to the model, simultaneously, the cloud samples the third-layer perturbation and performs the transmission of the second layer. 

\begin{figure}[tbp]
  \centering
  \includegraphics[width=\linewidth]{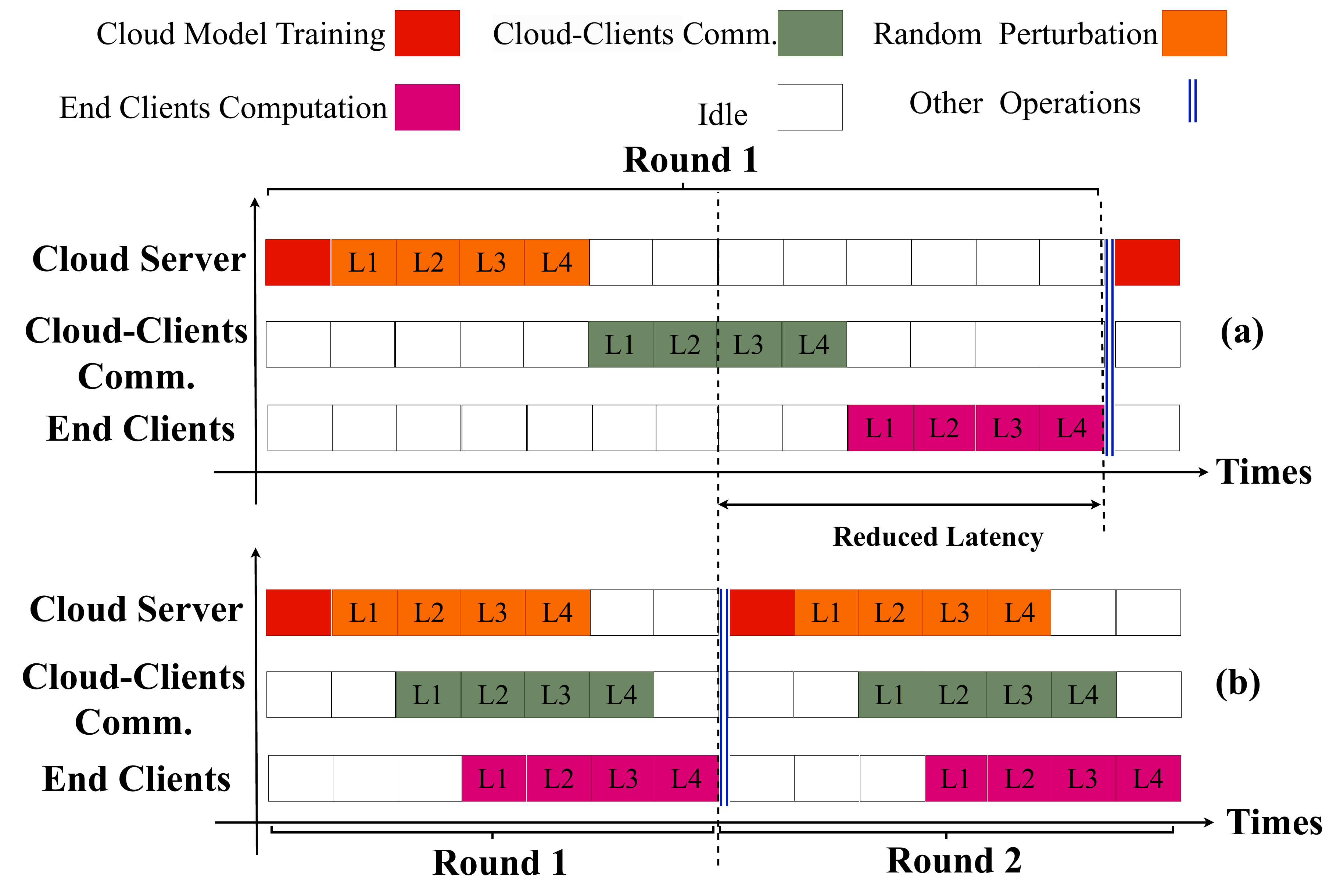}
  \caption{The comparison of Without and With System-level Pipeline in CooperLLM.}
  \label{fig:pipeline}
\end{figure}

As shown in Fig. \ref{fig:pipeline} (a), without the SPC, the transmission of the guided perturbation must wait until all layers of guided perturbations are fully sampled, and clients must wait for the complete transmission of all layers before computing the hybrid perturbations. This sequential processing introduces significant transmission and computation overhead. Fig. \ref{fig:pipeline} (b) shows the process of the SPC, and we can find that this layered parallelism effectively hides communication latency. 

Consider an LLM with $L$ layers. Without SPC optimization, the total delay for this transmission and computation stage is given by $T^0_{all} = T_{cs} + T_{t} + T_{cc}$. With SPC optimization, the total delay for this transmission and computation stage is given by $T^1_{all} = T_{cs} + \frac{1}{n}T_{cc}$, where $ T_{cs}, T_{t}, T_{cc}$ represent the guided perturbation sampling latency on cloud, the transmission latency, and the guided perturbation latency, respectively. The reduced latency:
\begin{align}
T^r_{all}&=T^0_{all}-T^1_{all}\\ \nonumber
&=T_{t} + \frac{L-1}{L}T_{cc}.
\end{align}
For example, consider LLaMA-2-7B with 32 layers, the reduced latency is $T_{t} + \frac{31}{32}T_{cc}$. The more layers the model has, the greater the latency reduction achieved. The more layers the model has, the higher the latency reduction rate achieved.

\noindent\textbf{Layer-wise Memory Reduction.} During the ZGR process, as shown in Figure \ref{fig:pipeline-1}, each client needs to store their local model, all layers’ local perturbations, guided perturbations, and forward gradient. With SPC, the storage requirement on clients is reduced from holding perturbations for all layers (both local and guided) to only keeping perturbations for a single layer at the same time. Moreover, with ADW (Sec. \ref{Sec:overview}), the forward gradients no longer need to be stored for all layers; instead, only a single scalar function loss is kept, whose memory footprint is negligible (a constant). 

Figure \ref{fig:pipeline-1} (a) shows that without SPC and ADW optimizations, the memory footprint for each client $Mem^0 = 4Mem_0$, where $Mem_0$ is the LLM model size. Figure \ref{fig:pipeline-1} (b) illustrates the that with SPC and ADW optimizations, the memory footprint for each client is $Mem^1 = (1+\frac{2}{L})Mem_0$. The reduced memory footprint:
\begin{align}
Mem_r &= Mem^0-Mem^1\\ \nonumber
&= (3-\frac{2}{L})Mem_0.
\end{align}
Note that the memory reported here refers to the average memory footprint, while the peak memory also includes the activation memory. Similarly, consider LLaMA-2-7B with 32 layers, the reduced memory footprint $Mem_r = (3-\frac{2}{32})Mem_0$. When storing the LLM in FP16 precision, the memory footprint of $Mem_0$ is about 13.6 GB. With our method, fine-tuning a LLaMA-2-7B LLM can reduce the client-side memory footprint by approximately 40 GB.

\subsection{Data Transmission Controller}\label{Sec:DTC}
\begin{figure}[tbp]
  \centering
  \includegraphics[width=\linewidth]{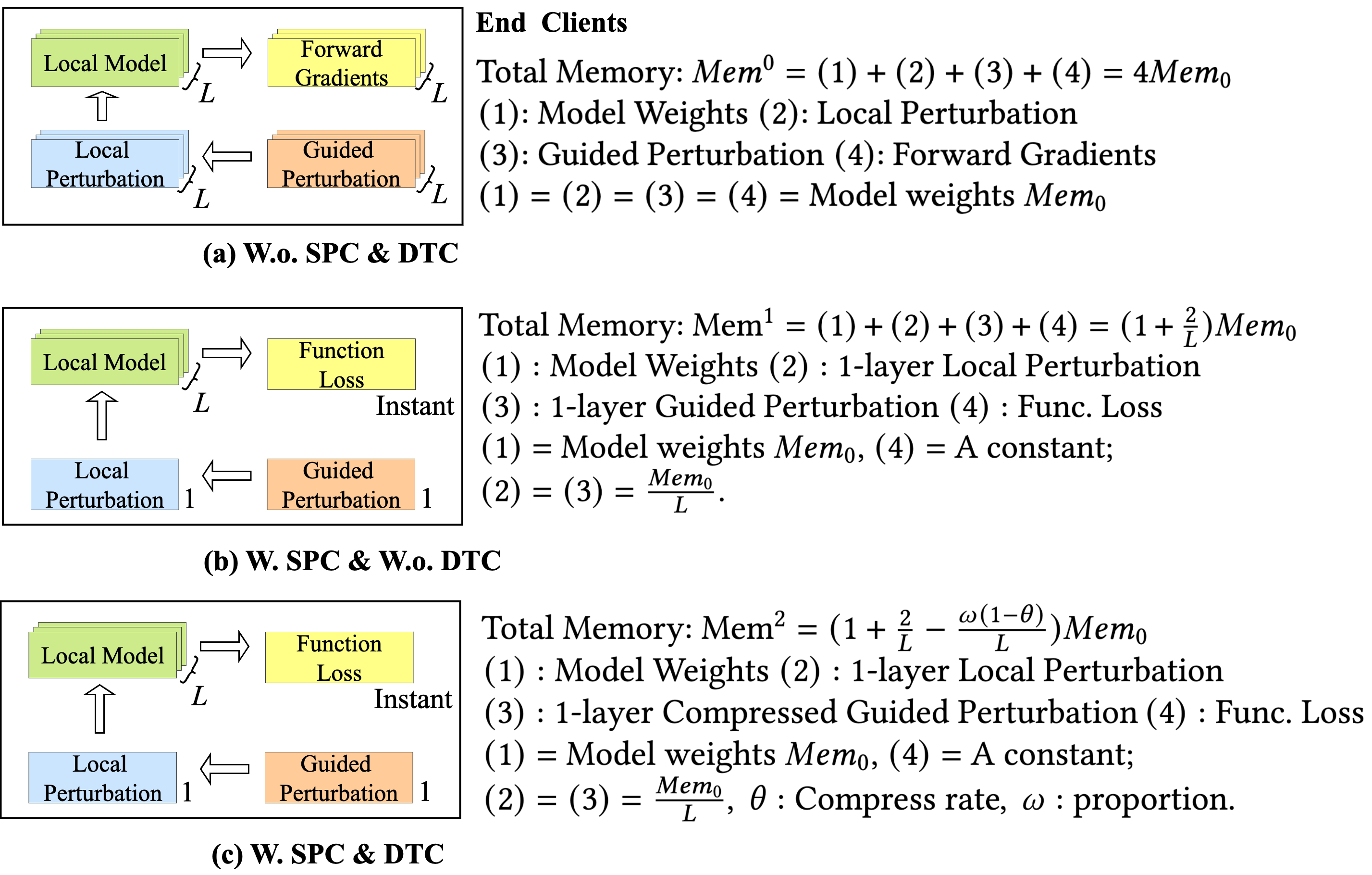}
  \caption{The Memory Footprint Comparison with/without SPC\&DTC.}
  \label{fig:pipeline-1}
\end{figure}

DTC builds upon SPC by further optimizing communication to reduce latency. In SPC, guided perturbations are transmitted layer by layer in a pipelined manner. However, for the communication latency to be perfectly hidden, it must remain smaller than the computation latency. Note that in this setting, it is sufficient to satisfy the computation latency on the client side, since sampling on the cloud side can be completed quickly and does not depend on either transmission or client-side computation. To address this, we adopt data quantization and compression to reduce the transmitted data volume, thereby decreasing communication latency. We leverage a robust compression method \cite{chmiel2020robust}, which has been widely adopted in recent works on efficient large language models \cite{zhao2024alisa, sun2025breaking}. Specifically, we adopt a quantization scheme that maps guided perturbation into $b$-bit integers for storage in memory, and subsequently restores them to the former precision during computation. The procedure is defined as follows:
\begin{equation}
    x_{\text{quant}} = \text{round}\left(\frac{1}{\lambda}x + \delta \right), 
    \quad 
    x = \lambda \left(x_{\text{quant}} - \delta \right),
\end{equation}
where the scaling factor is given by $\lambda = \frac{\max - \min}{2^b - 1}$ and the zero point is determined by $\delta = \text{round}\left(\frac{-2^b}{\max - \min}\right)$. In CooperLLM, guided perturbations are compressed to INT4 precision for transmission efficiency, and later decompressed back to FP16 precision before being applied to the model. However, decompression on the client also incurs additional computation latency, raising the challenge of how to control the amount of compressed data. We assume a compression proportion $\omega \in (0,1)$ and a compression ratio $\theta \in (0,1)$. The transmission latency per layer is given by
\begin{equation}
T_t = \frac{D_0}{B_0},    
\end{equation}
while the computation latency on the client side is
\begin{equation}
T_{cc} + T(D_0 \cdot \omega),    
\end{equation}
where $D_0$ denotes the data size, $B_0$ the available bandwidth, and $T(D_0 \cdot \omega)$ the decompression time, which can be obtained offline. To ensure that the transmission delay can be hidden within the computation delay, it suffices to satisfy
\begin{equation}
T_t = \frac{D_0}{B_0} \leq T_{cc} + T(D_0 \cdot \omega).    
\end{equation}
Hence, the proportion $\omega$ can be configured adaptively based on the above parameters. Figure \ref{fig:pipeline-1} (c) shows that when using SPC, the client memory usage is $Mem^1 = (1+\frac{2}{L})Mem_0$. With DTC, the memory usage is $Mem^2 = (1+\frac{2}{L}-\frac{\omega(1-\theta)}{L})Mem_0$.

\section{Implementation settings}
\textbf{Implementation}. We implement CooperLLM on top of \cite{lin2021fednlp}, a recent federated learning framework for NLP tasks. We chose this work because it supports a wide range of NLP task structures and offers good scalability in federated architectures. The BP-based fine-tuning on the cloud is implemented by ourselves. We use FedSGD \cite{yuan2020federated} to aggregate in our FL system. For the preliminary experiments in Sec. \ref{Bg_Mo}, we implement them based on Hugging Face Transformers \cite{wolf2019huggingface} and Pytorch \cite{paszke2019pytorch}. We conduct these experiments on a cloud platform in a similar setting to the prior works \cite{lai2021oort,li2022pyramidfl,xu2024fwdllm,li2021hermes} on $4 \times$  NVIDIA A40 GPU with 56 GB GPU memory for each. We partition the GPU resources, following the CooperLLM framework, into three tiers: cloud, edge servers, and clients. Meanwhile, we also conduct on-device training on Jetson TX2 (8GB) \cite{nvidia-jetson-tx2} and Jetson Orin NX (16GB) \cite{nvidia-jetson-orin} (See TABLE \ref{tab:spec} for details) to evaluate memory usage and other system metrics.

\begin{table}[tbph]
  \centering
  \caption{Specifications of tested devices. (Nvid.:Nvidia)}
  \label{tab:spec}
  \begin{tabular}{lccc}
    \toprule
    \textbf{Device} & \textbf{Mem} & \textbf{GPU} & \textbf{CPU} \\
    \midrule
    Nvid. Jetson TX2 & 8\,GB & 256 CUDA Cores & A57\\
    Nvid. Jetson Orin NX & 16\,GB & 1024 CUDA Cores & A78\\
    \bottomrule
  \end{tabular}
\end{table}

\begin{figure*}[htbp]
    \centering
    \begin{subfigure}{0.33\textwidth}
        \includegraphics[width=\linewidth]{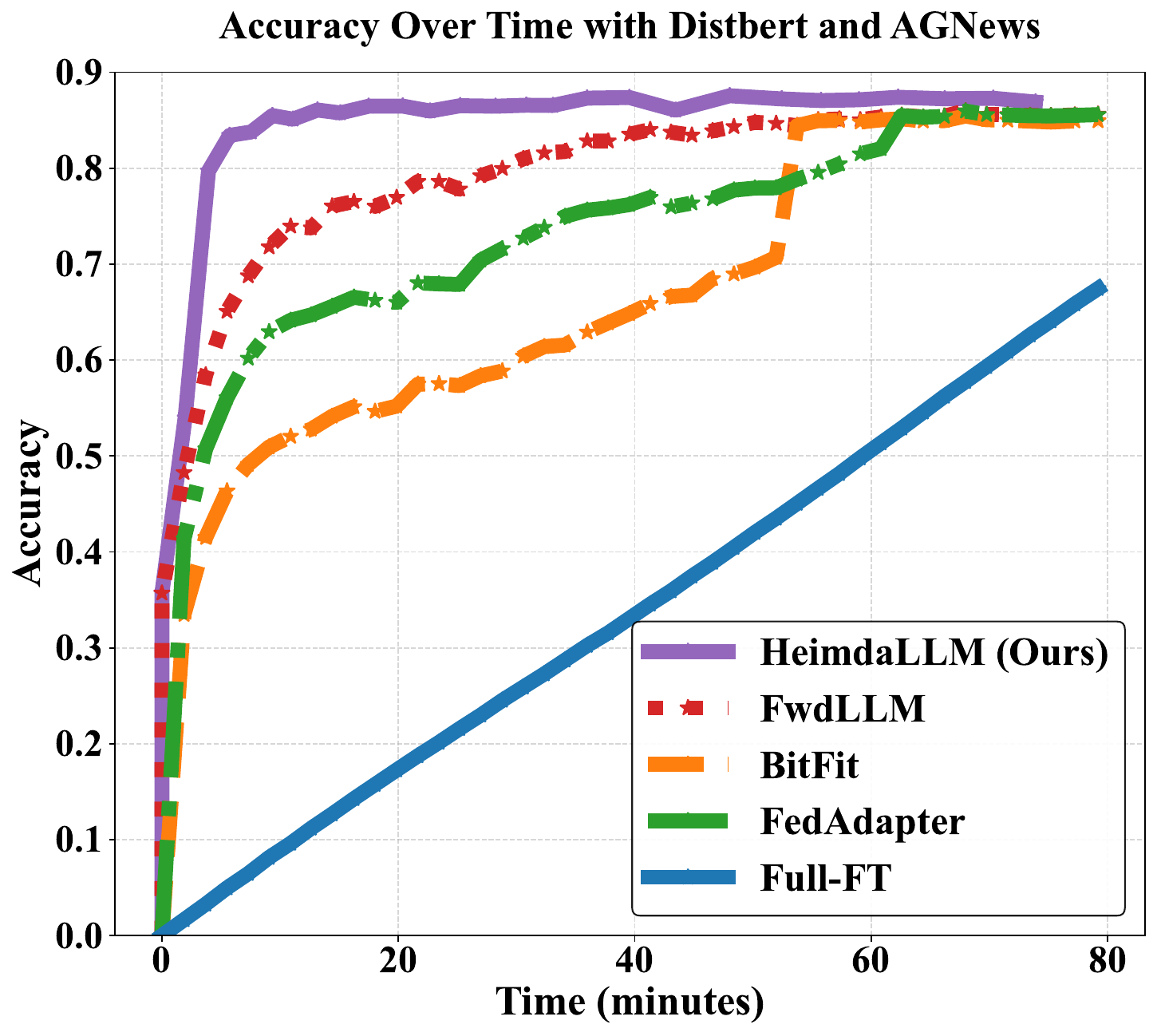}
        \caption{Accuracy Comparison with DistilBERT-base and AGNEWS.}
        \label{fig:subfig1}
    \end{subfigure}
    \hfill
    \begin{subfigure}{0.33\textwidth}
        \includegraphics[width=\linewidth]{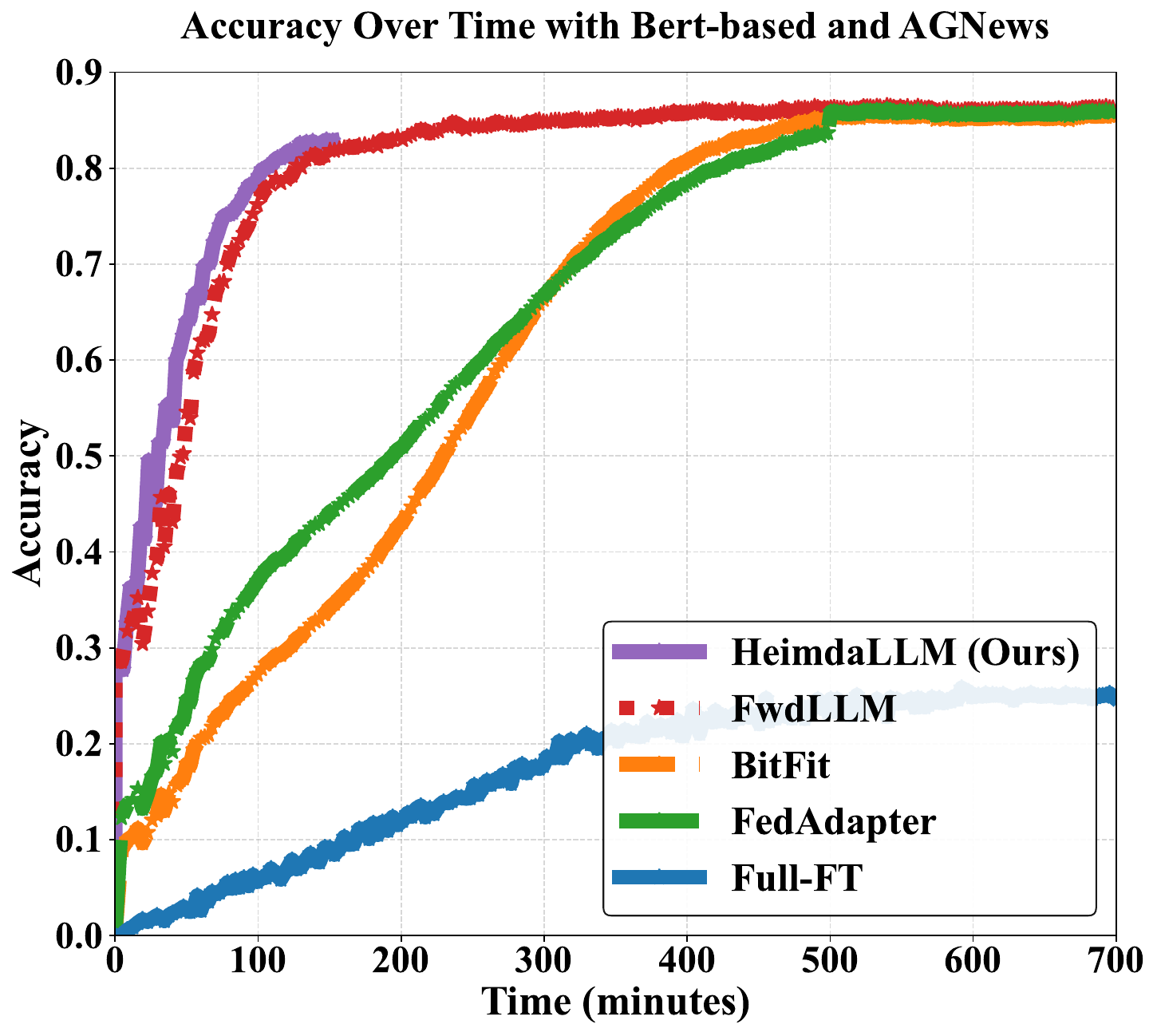}
        \caption{Accuracy Comparison with BERT-base and AGNEWS.}
        \label{fig:subfig2}
    \end{subfigure}
    \hfill
    \begin{subfigure}{0.33\textwidth}
        \includegraphics[width=\linewidth]{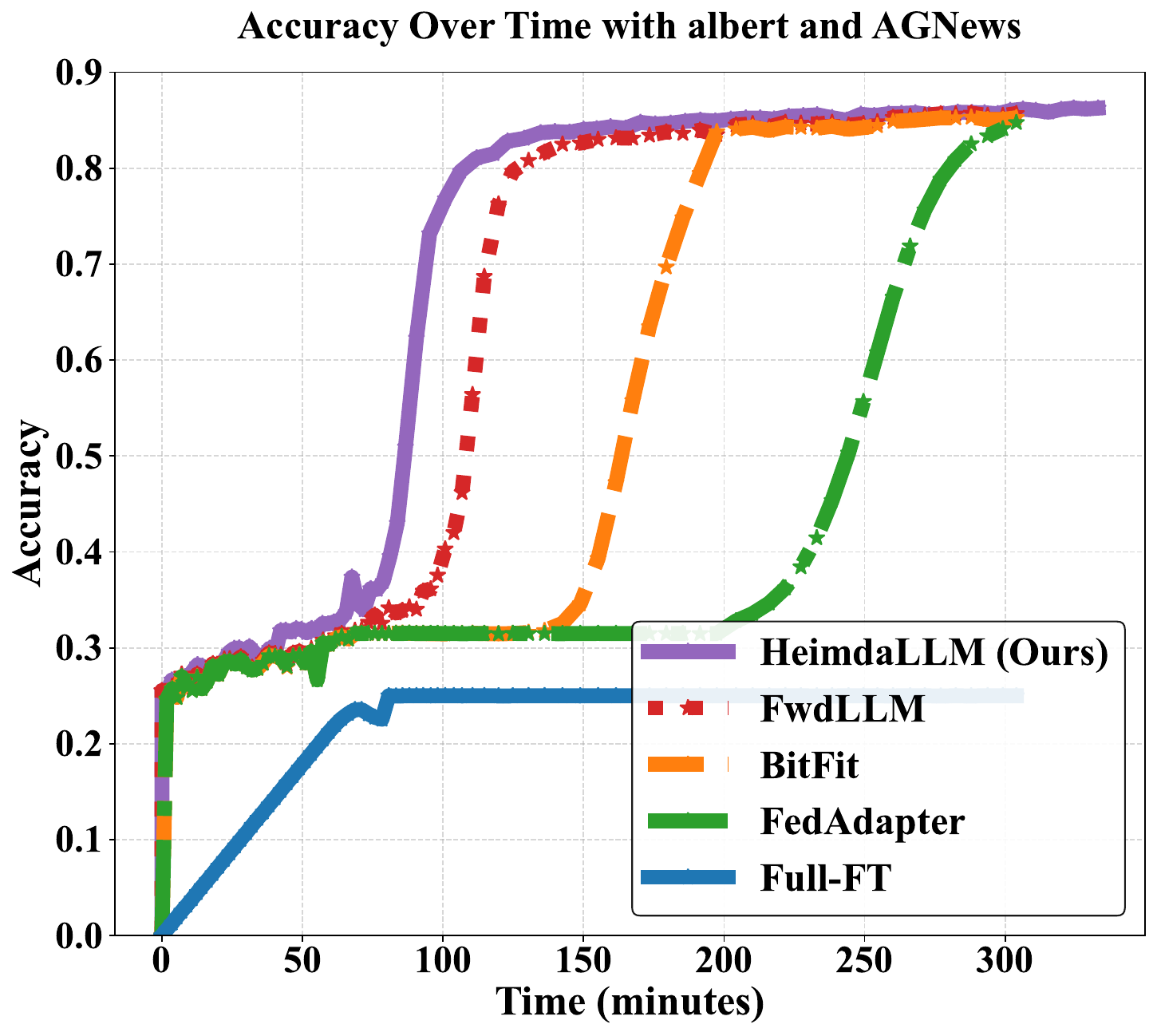}
        \caption{Accuracy Comparison with ALBERT-base and AGNEWS.}
        \label{fig:subfig3}
    \end{subfigure}

    \begin{subfigure}{0.33\textwidth}
        \includegraphics[width=\linewidth]{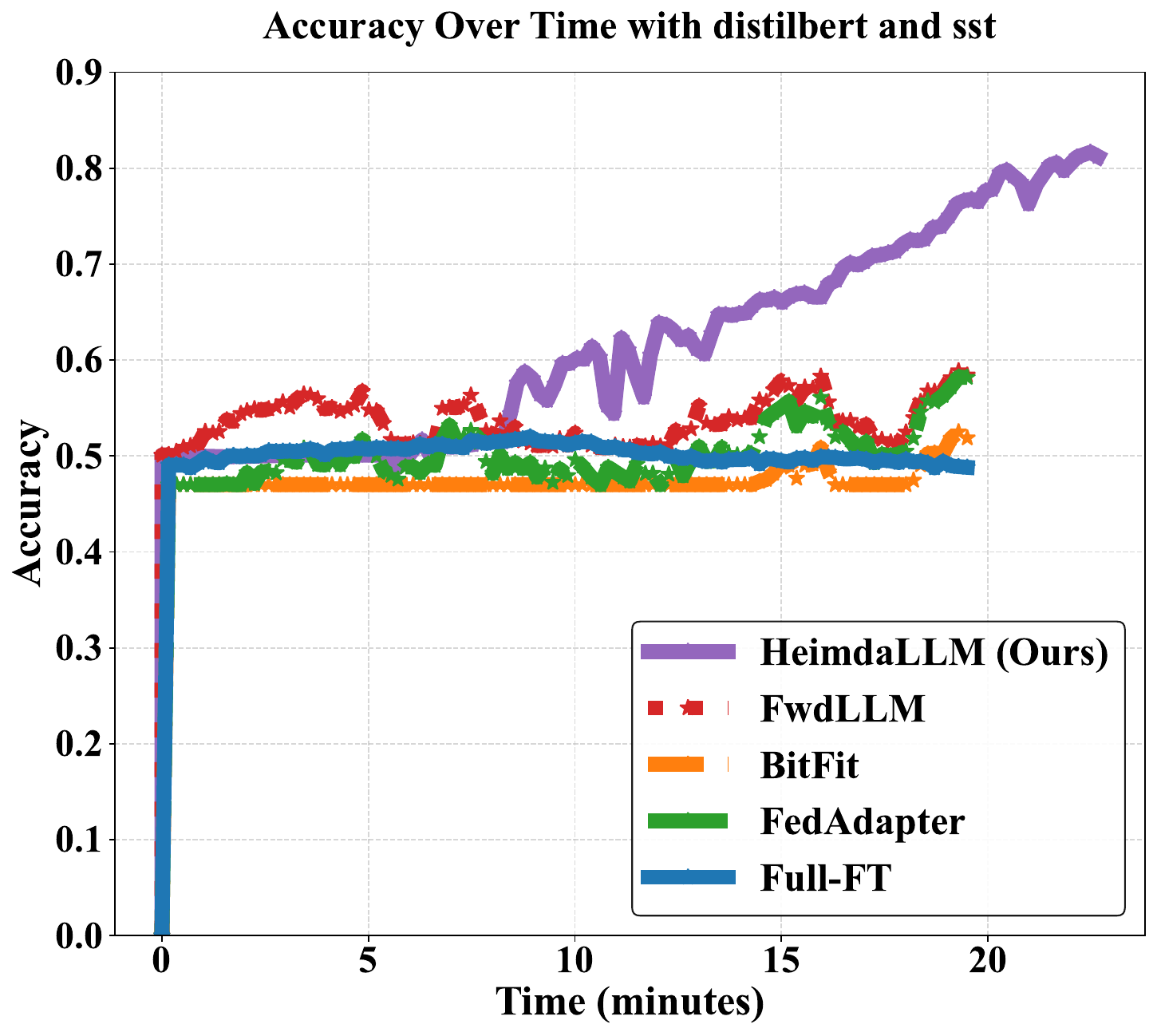}
        \caption{Accuracy Comparison with DistilBERT-base and SST-2.}
        \label{fig:subfig4}
    \end{subfigure}
    \hfill
    \begin{subfigure}{0.33\textwidth}
        \includegraphics[width=\linewidth]{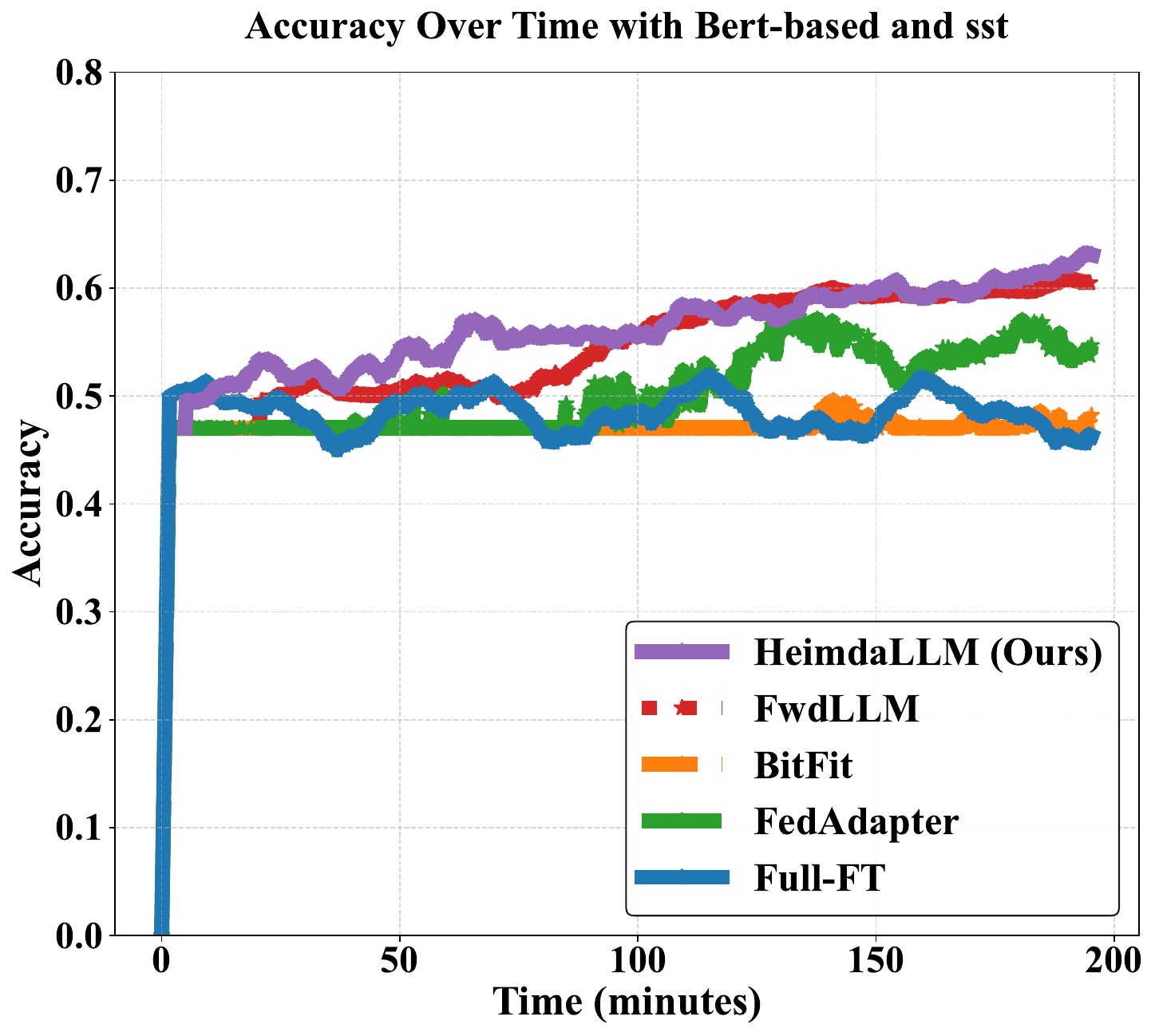}
        \caption{Accuracy Comparison with BERT-base and SST-2.}
        \label{fig:subfig5}
    \end{subfigure}
    \hfill
    \begin{subfigure}{0.33\textwidth}
        \includegraphics[width=\linewidth]{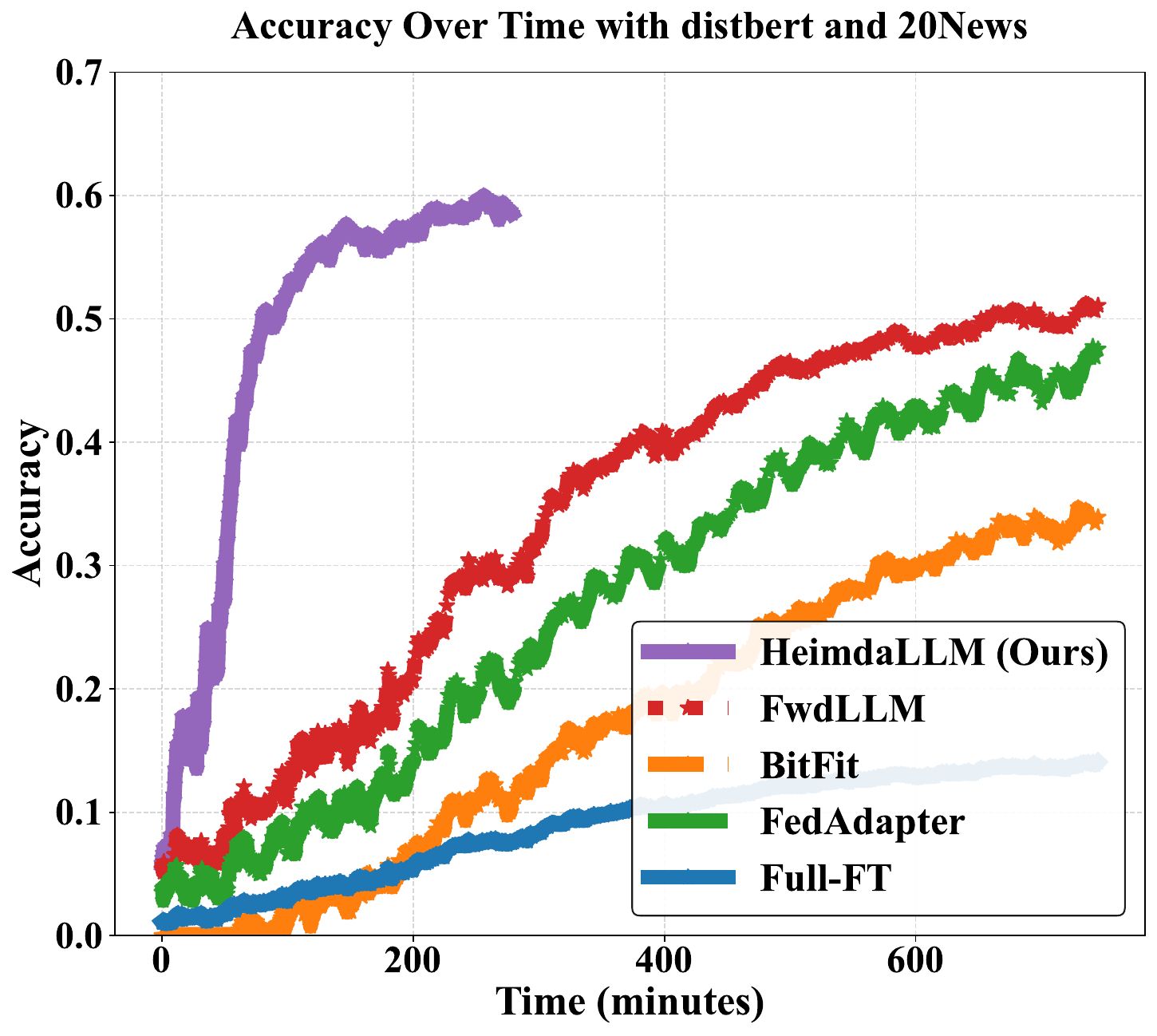}
        \caption{Accuracy Comparison with DistilBERT-base and 20News.}
        \label{fig:subfig6}
    \end{subfigure}


    \caption{Accuracy comparisons with different models and different datasets.}
    \label{fig:overall_performence}
\end{figure*}

\noindent\textbf{Models.} We evaluate CooperLLM using five models: three BERT-like encoder-only architectures: (1) DistilBERT-base \cite{sanh2019distilbert} wit 66M parameters, (2) ALBERT-base \cite{lan2019albert} with 12M parameters, (3) BERT-based \cite{devlin2019bert} with 110M parameters, (4) RoBERTa-based \cite{liu2019roberta} with 340M parameters for memory footprint test; one model from the LLaMA series: Llama-2-7B \cite{touvron2023llama} with 7B parameters. In order to fill the whole LLaMA model in memory, we use the quantized model of Llama-2-7B, the same as \cite{xu2024fwdllm} for inference. All above models are obtained from the HuggingFace \cite{wolf2019huggingface}. 

\noindent\textbf{Datasets.} We evaluate CooperLLM on four representative NLP datasets: (1) SST-2 \cite{zhang2015character}, a sentiment classification dataset based on restaurant reviews, with 280K training and 19K testing samples for each polarity; (2) AGNEWS \cite{zhang2015character}, a news classification benchmark with 4 classes, each containing 30K training and 1.9K testing samples; (3) 20News \cite{lang1995newsweeder}, a widely-used text classification benchmark that encompasses 20 diverse newsgroups, each representing a distinct topic such as sports, technology, and politics. It contains approximately 20,000 documents, with roughly 1,000 articles per category; and (4) LongAlign \cite{bai2024longalign},  a specialized resource designed to enhance the alignment of long-text descriptions with generated content, where the context length can reach up to 64K. For the memory overhead, we sample from LongAlign with different context lengths. 

\noindent \textbf{Metrics.} We use the time-to-accuracy metric (\%) to test the accuracy performance of CooperLLM and baselines, and we use top-1 accuracy to evaluate the models’ performance across different datasets. To test the convergence latency of CooperLLM and baselines, we use time metric (s). As for the memory overhead testing, we use the memory metric (MB/GB) to compare CooperLLM with baselines. Meanwhile, to compare the acceleration effect of CooperLLM with the baselines, we also report SpeedUp ($\times$) as the metric.

\noindent \textbf{Baselines.} 
We evaluate our method, CooperLLM, against four baselines: (1) Federated Full-Tuning \cite{lin2021fednlp}: In this baseline, the FL system updates all model parameters during training. (2) Bitfit \cite{zaken2021bitfit}: Instead of updating all parameters, this approach fine-tunes only the bias terms of each layer in the LLM. (3) FedAdapter \cite{cai2023efficient}: It combines adapter tuning with selective layer freezing and further adopts a progressive training strategy that can automatically determine the most suitable adapter configuration. (4) FwdLLM \cite{xu2024fwdllm}: It is a BP-free federated fine-tuning protocol that replaces backpropagation with perturbed inferences, enabling memory-efficient and scalable FedLLM on resource-constrained devices.
The baselines (1)–(3) are BP-based, while (4) FwdLLM is BP-free. Notably, FwdLLM represents the state-of-the-art BP-free approach for addressing memory constraints in federated learning. These baselines feature algorithm designs that reduce memory usage while maintaining accuracy. Building on such algorithmic foundations, CooperLLM additionally incorporates system-level design, and these baselines provide a comparison in terms of both algorithmic flexibility and system-level efficiency.











\noindent \textbf{FL Settings.} In all baselines, 80 clients are selected in each round. For CooperLLM, we configure one cloud server, one edge server, and 80 end clients. Both CooperLLM and all baselines adopt the same FL hyperparameter settings to ensure fairness. The network bandwidth is set to 10 Mbps, following prior work \cite{cai2023efficient,lin2021fednlp}. The number of local epochs is set to 1 for both clients and the cloud, with a learning rate of 0.1. The mini-batch size is 8 on the client side and 16 on the cloud side. In all experiments, unless otherwise stated, we set the parameter $\alpha$ to 0.5. We use FedSGD \cite{yuan2020federated} in the FL system. 


\section{Evaluation}
\subsection{Overall performance.}
In this subsection, we compare the accuracy of CooperLLM and the baselines across different models (DistilBERT, ALBERT-base, and BERT-base) and datasets (SST-2, AGNEWS, 20NEWS) according to the configurations described in the FL settings with batch size = 8.
\begin{table*}[ht]
    \centering
    \caption{Test results (top-1 accuracy and speedup) on AGNews, SST-2, 20News}
    \label{tab:results}
    \begin{tabular}{@{}lcccccccccc@{}}
        \toprule
        Dataset & \multicolumn{2}{c}{Full-FT} & \multicolumn{2}{c}{BitFit} & \multicolumn{2}{c}{FedAdapter} & \multicolumn{2}{c}{FwdLLM} & \multicolumn{2}{c}{CooperLLM (Ours)} \\
        \cmidrule(lr){2-3} \cmidrule(lr){4-5} \cmidrule(lr){6-7} \cmidrule(lr){8-9} \cmidrule(lr){10-11}
        & Speedup & Acc (\%) & Speedup & Acc (\%) & Speedup & Acc (\%) & Speedup & Acc (\%) & Speedup & Acc (\%) \\
        \midrule
        AGNews & 1$\times$ & 67.45 & 1.45$\times$ & 84.94 & 1.77$\times$ & 85.58 & 2.66$\times$ & 85.61 & \textbf{8.88$\times$} & \textbf{87.55} \\
        SST-2 & 1$\times$ & 48.82 & 1.03$\times$ & 52.51 & 1.04$\times$ & 58.22 & 1.12$\times$& 58.86 & \textbf{1.81$\times$} & \textbf{81.65} \\
        20News & 1$\times$ & 14.07 & 1.06$\times$ & 33.92 & 1.14$\times$ & 47.55 & 1.6$\times$ & 51.10 & \textbf{4.4$\times$} & \textbf{59.98} \\
        \bottomrule
    \end{tabular}
\end{table*}
\begin{figure*}[htbp]
    \centering
    \begin{subfigure}{0.49\textwidth}
        \includegraphics[width=\linewidth]{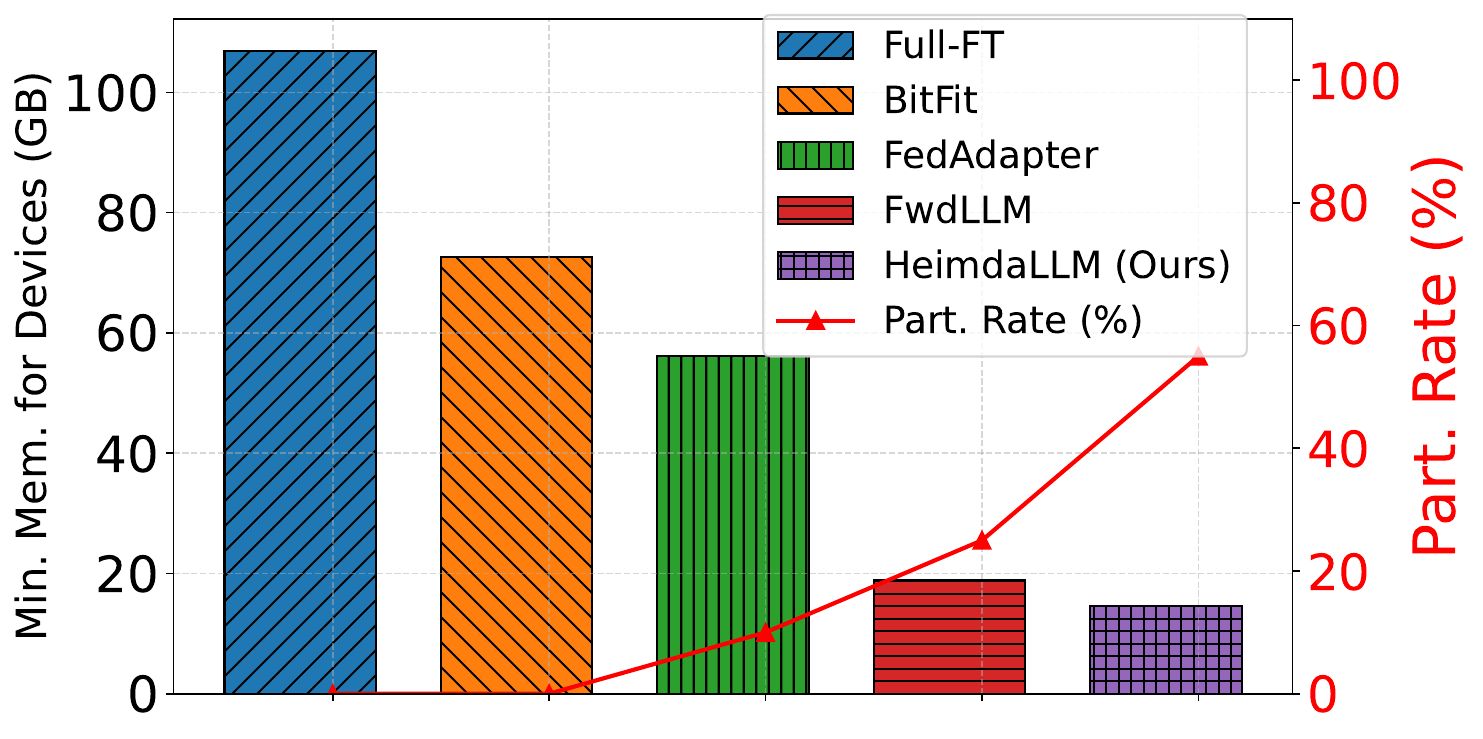}
        \caption{Memory Footprint and Participant Rate Comparisons on Llama-7B model.}
        \label{fig:subfig21}
    \end{subfigure}
    \hfill
    \begin{subfigure}{0.49\textwidth}
        \includegraphics[width=\linewidth]{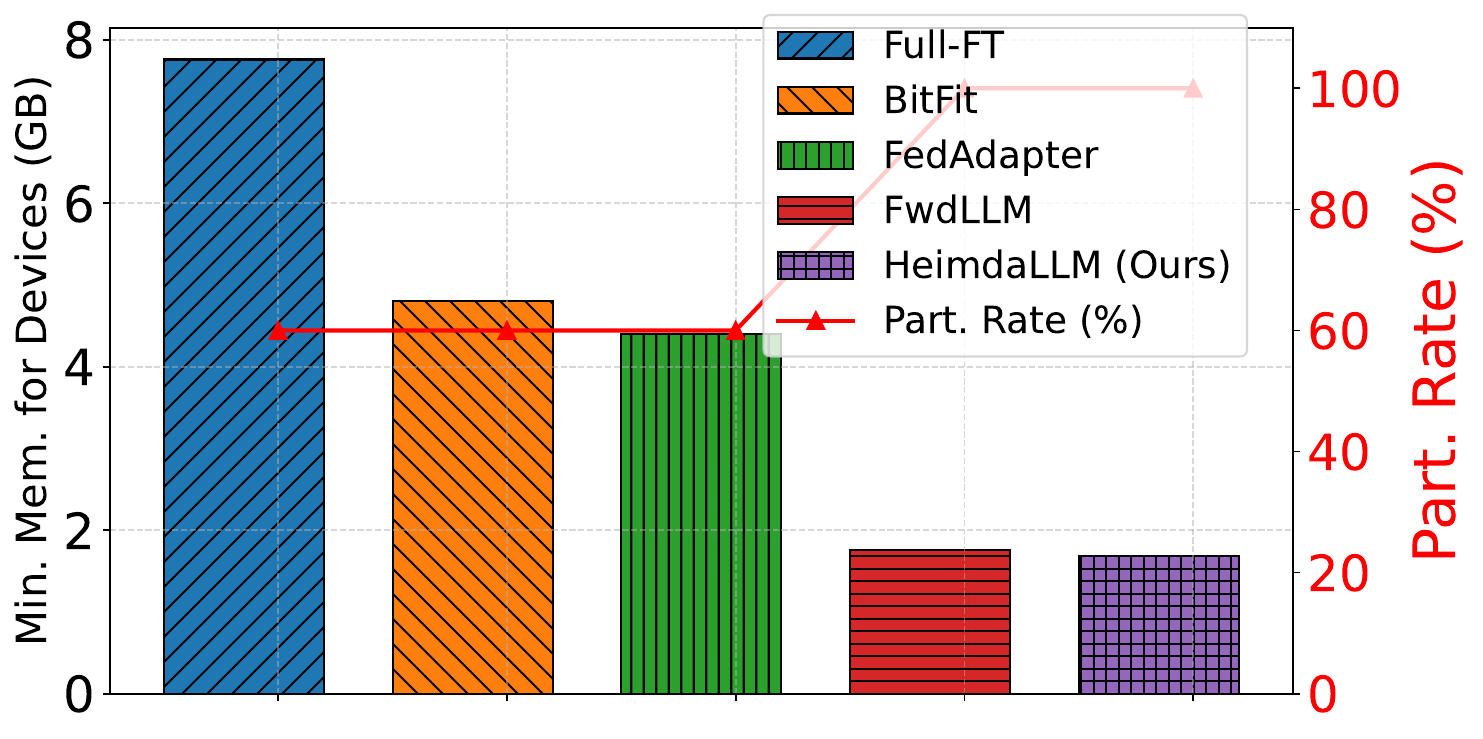}
        \caption{Memory Footprint and Participant Rate Comparisons on RoBERTa-Based Model.}
        \label{fig:subfig22}
    \end{subfigure}

    \caption{Memory Footprint and Participant Rate Comparisons between CooperLLM and Baselines.}
    \label{fig:Mem_performence}
\end{figure*}

\noindent\textbf{Superior Accuracy.} Figure \ref{fig:overall_performence} shows the accuracy comparison results. Overall, across different models and datasets, CooperLLM consistently achieves higher accuracy compared to the baselines, particularly demonstrating superior accuracy on the SST-2 and complex 20News datasets with the DistilBERT-based model. 

Specifically, on the AGNews dataset with the DistilBERT-based model (figure \ref{fig:subfig1}), CooperLLM outperforms FwdLLM, BitFit, and FedAdapter by approximately 2 percentage points in peak accuracy at convergence, and surpasses the Full-FT method by approximately 17.8 percentage points. On the AGNews dataset with the BERT-based model and BERT-based model (figure \ref{fig:subfig2} and figure \ref{fig:subfig3}), CooperLLM achieves approximately $2\sim 3$ percentage points higher accuracy at convergence compared to FwdLLM, and reaches a similar final accuracy to the PEFT methods BitFit and FedAdapter. However, CooperLLM exhibits a significantly faster convergence speed. On the SST-2 dataset with the DistilBERT-based model (figure \ref{fig:subfig4}), the baselines generally fluctuate around 50\% accuracy, with FwdLLM performing slightly better than the others. However, CooperLLM outperforms all these baselines by approximately $21 \sim 24$ percentage points in accuracy. On the SST-2 dataset with the BERT-based model (figure \ref{fig:subfig5}), the baselines also generally fluctuate around 50\% accuracy, with FwdLLM performing slightly better than the others. However, CooperLLM outperforms all these baselines by approximately $17 \sim 19$ percentage points in accuracy.

\noindent\textit{For complex tasks, 20News dataset}, as shown in figure \ref{fig:subfig6}, we observe that CooperLLM significantly outperforms the baselines in LLM fine-tuning for such complex tasks. For instance, compared with the recent method FwdLLM, our approach achieves approximately 10 percentage points higher accuracy. For other PEFT methods such as BitFit and FedAdapter, CooperLLM achieves approximately 14.3 and 26 percentage points higher accuracy, respectively. Remarkably, it even surpasses the full-parameter fine-tuning method (Full-FT) by approximately 47 percentage points.

\noindent\textbf{Shorter convergence time.} 
For CooperLLM and different baselines, we evaluated top-1 accuracy and convergence speed on the DistilBERT-based model across three datasets: AGNews, SST-2, and 20News, as shown in Table \ref{tab:results}. We set the batch size as 8.

Overall, CooperLLM converges the fastest, achieving a speedup of $4.4\times \sim 8.8\times$ compared with the baselines. Figure \ref{fig:overall_performence} also shows CooperLLM achieves higher top-1 accuracy than all the baselines. 

Specifically, taking Full-FT as the baseline unit of speed, we evaluated the acceleration of different baselines and CooperLLM. On the AGNews dataset, BitFit, FedAdapter, and FwdLLM can achieve convergence speedups of $1.45\times$, $1.77\times$, and $2.66\times$, respectively. CooperLLM, on the other hand, achieves an 8.88× speedup, representing improvements of 612.4\%, 501.7\%, and 333.8\% over the baselines, respectively.
On the SST-2 dataset, BitFit, FedAdapter, and FwdLLM can achieve convergence speedups of $1.03\times$, $1.04\times$, and $1.12\times$, respectively. CooperLLM, on the other hand, achieves $1.81\times$ speedup, representing improvements of 175.7\%, 174.0\%, and 161.6\% over the baselines, respectively. 
On the 20News dataset, BitFit, FedAdapter, and FwdLLM can achieve convergence speedups of $1.06×$, $1.14×$, and $1.6×$, respectively. CooperLLM, on the other hand, achieves $4.4\times$ speedup, representing improvements of 415.1\%, 385.9\%, and 275\% over the baselines, respectively. 

In terms of top-1 accuracy, CooperLLM achieves 87.55\%, 81.65\%, and 59.98\% on AGNews, SST-2, and 20News, respectively.
\begin{figure}[tbp]
  \centering
  \includegraphics[width=\linewidth]{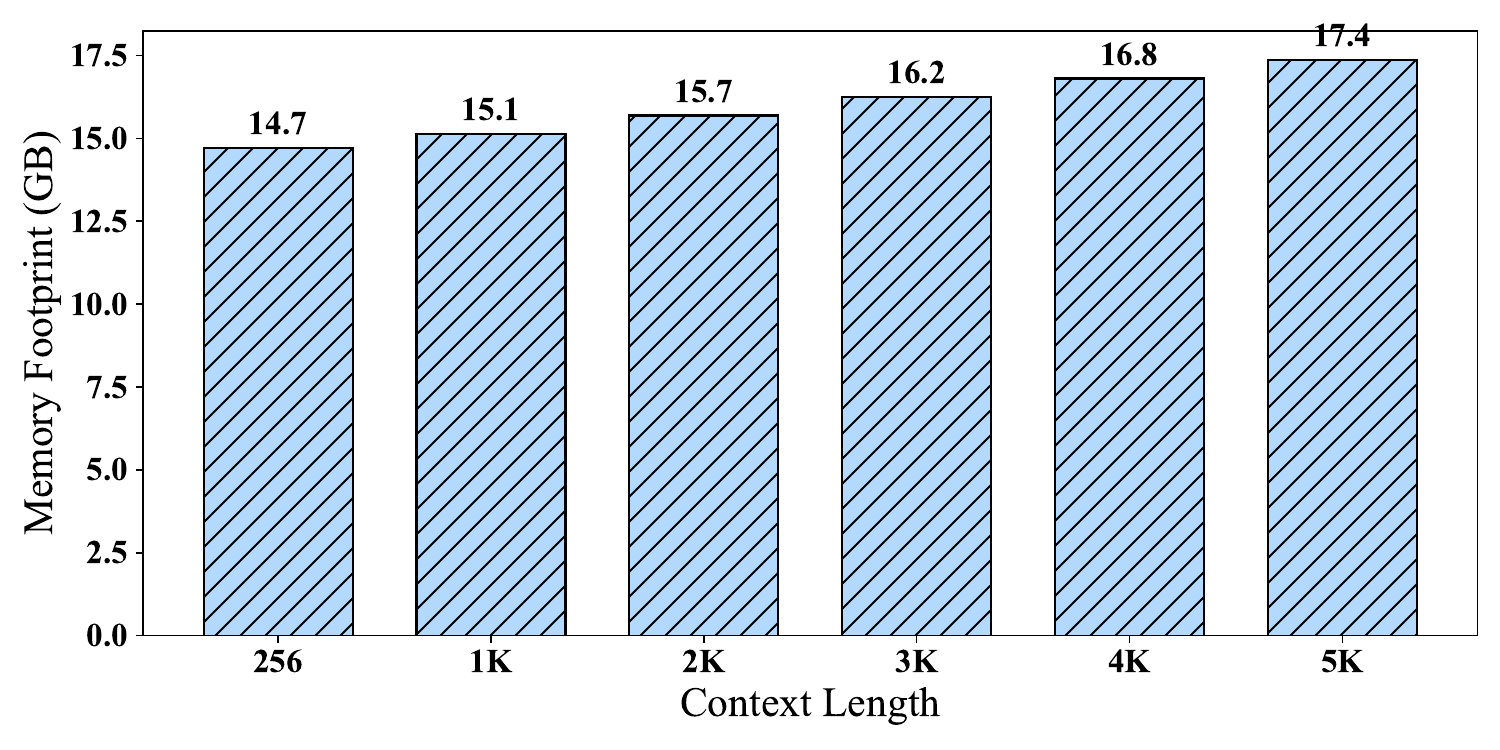}
  \caption{The Memory Footprint with Different Context Length.}
  \label{fig:mem_sen}
\end{figure}

\subsection{System Overhead.}
In this subsection, we adopt the simulation setup described in Sec. \ref{Sec:Mem_Break} with batch size = 1 and context length = 1k. Meanwhile, we also evaluate the memory footprint of CooperLLM under the Jetson devices configuration (Table \ref{tab:spec}). Additionally, we analyze the end-to-end latency of CooperLLM.

\noindent\textbf{Memory footprint.} Figure \ref{fig:Mem_performence} presents the results of our memory footprint and participation rate calculations across different models (Llama-7B and RoBERTa-based models). Specifically, CooperLLM and FwdLLM exhibit comparable and slightly lower memory usage, as we offload the memory footprint of gradients to the Edge Server. Compared with the other baselines, the reduction in memory usage is significant. On Llama-7B, CooperLLM reduces memory consumption by 86.37\%, 79.95\%, and 74.05\% relative to the baselines; on the RoBERTa-based model, the reductions are 78.22\%, 64.79\%, and 61.59\%, respectively. Regarding participation, PEFT indeed reduces part of the memory footprint compared to full-parameter fine-tuning methods. In contrast, CooperLLM and FwdLLM, which are ZOO-based methods, reduce the majority of the memory footprint compared to both full-parameter fine-tuning and PEFT. Due to the system-level design, our memory usage is further reduced by 22.6\% and 4\% on the Llama-7B and RoBERTa-base models, respectively, compared to FwdLLM.

\noindent\textbf{End-to-end data transmission latency.} 
Compared with conventional LLM fine-tuning frameworks, CooperLLM introduces an additional transmission delay caused by the guided perturbation transfer from the cloud to end devices. We optimize this delay using SPC and DTC, resulting in an added latency equivalent to transmitting less than one model layer’s data. For instance, with Llama-7B at a compression ratio of 0.8 and rectification performed every 10 rounds, the average additional data transmission delay per round is only 13.48 seconds, which is negligible compared to the training time of several tens of minutes per round. 

Moreover, in the ZGR design, each end client only needs to upload a constant-sized loss per round, rather than the full gradient, which greatly reduces end-to-end latency. Note that the only additional overhead comes from transmitting the cloud-to-server guided perturbations, but this delay can be entirely hidden within the training process, so it does not introduce any extra end-to-end latency.

\subsection{Sensitive Analysis and Ablation Study.}
In this subsection, we conduct a sensitivity analysis on context length and further perform an individual design breakdown to evaluate the impact of each technique on memory usage and end-to-end latency with Llama-7B. 

\noindent\textbf{Context Length Sensitive Analysis.} Since the memory footprint of ZOO-based LLM fine-tuning methods also varies with the context length, we propose the SPC\&DTC designs for optimization. Therefore, we further analyze the impact of context length on CooperLLM. We evaluated the memory footprint for long-context processing on the LongAlign dataset using the Llama-7B model. Figure \ref{fig:mem_sen} shows that the memory footprints increase with the growth of context length. We can adopt a method similar to manual checkpointing in GPipe \cite{huang2019gpipe}, saving only a few layers of activations while recomputing others on the fly to free up space. In this way, the impact of long-context data on CooperLLM’s memory footprint is significantly mitigated in a 1K-context gap. 
\begin{figure}[tbp]
  \centering
  \begin{subfigure}[b]{\linewidth}
    \centering
    \includegraphics[width=\linewidth]{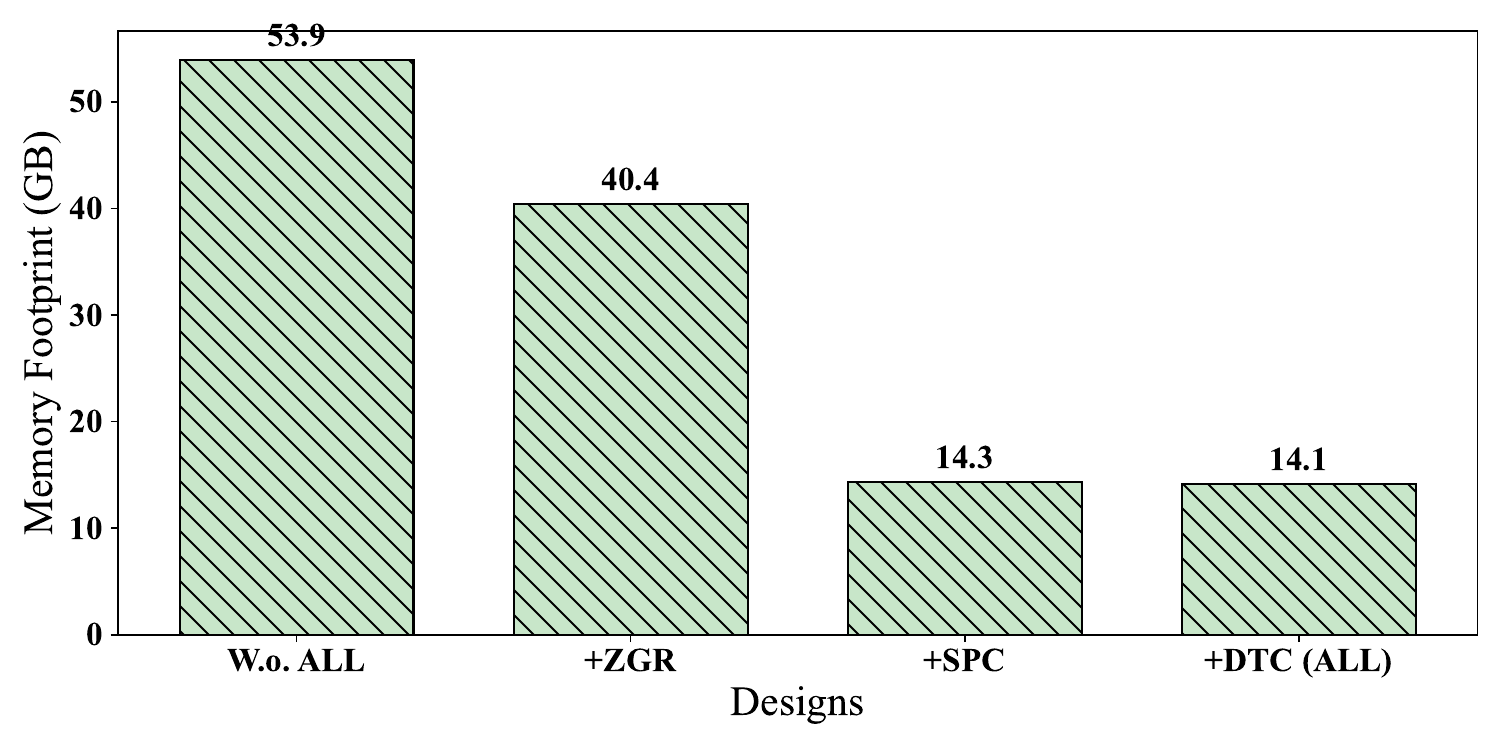}
    \caption{The Memory Breakdown on Individual Designs.}
    \label{fig:mem_indivi}
  \end{subfigure}
  
  \vspace{0.5cm} 
  
  \begin{subfigure}[b]{\linewidth}
    \centering
    \includegraphics[width=\linewidth]{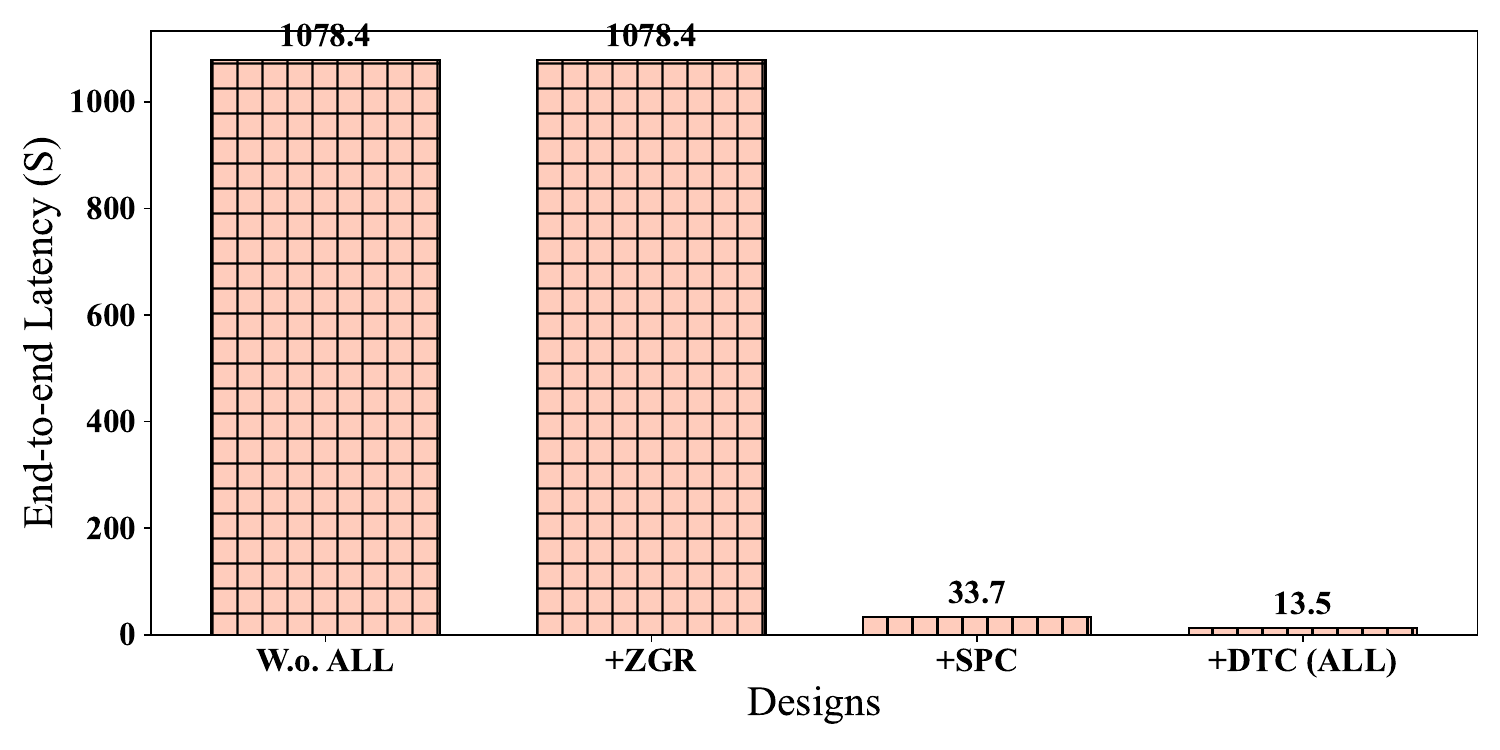}
    \caption{The Latency Breakdown on Individual Designs.}
    \label{fig:Latency_indivi}
  \end{subfigure}
  \caption{Performance breakdown analysis: (a) memory footprint and (b) latency across individual designs.}
  \label{fig:performance_breakdown}
\end{figure}

\noindent\textbf{Individual Designs Breakdown.} Figure \ref{fig:performance_breakdown} illustrates the contributions of different designs to memory consumption and end-to-end latency. With CooperLLM’s three design components, we achieve significant reductions in both overheads. Specifically, as shown in Figure \ref{fig:mem_indivi}, ZGR transfers the gradient memory footprint to the server instead of the end devices, thereby reducing the memory footprint. SPC ensures that only one layer of perturbations needs to be stored in device memory instead of all layers of LLM, while DTC compresses perturbations during inter-layer transmission, reducing the on-device memory footprint further. Together, these three designs complement each other, making LLM fine-tuning feasible on resource-constrained end devices.


\section{Related Work}
\noindent\textbf{Federated Fine-tuning for LLMs.} Federated fine-tuning for LLMs (FedLLM) has become the primary paradigm for customizing models for downstream tasks while preserving data privacy \cite{cai2023federated,bai2024federated,lin2021fednlp,cai2023efficient,wu2025survey,babakniya2023slora,che2023federated}. However, FedLLM faces the memory wall on resource-constrained mobile devices. To further address this challenge, numerous efforts have been made. For instance, PEFT methods (e.g., LoRA \cite{hu2021lora,dettmers2024qlora,chenlonglora,kuang2024federatedscope}, Adapter \cite{pfeiffer2020adapterhub,cai2023efficient,ghiasvand2024communication,kim2023client}) reduce memory consumption by decreasing the number of trainable parameters. 

The work \cite{dettmers2024qlora} proposes QLoRA, an efficient fine-tuning approach that enables 65B-parameter LLMs to be fine-tuned on a single 48GB GPU while preserving full 16-bit performance. It introduces 4-bit NormalFloat quantization, double quantization, and paged optimizers to minimize memory usage without degrading accuracy. The work \cite{kuang2024federatedscope} develops a comprehensive end-to-end pipeline for federated LLM fine-tuning and introduces off-site-tuning strategies to alleviate communication and computation overheads. The work \cite{cai2023efficient} introduces a progressive adapter tuning approach that leverages continuous device profiling to dynamically adjust adapter configurations across clients, thereby enhancing efficiency while maintaining accuracy. The work \cite{ghiasvand2024communication} employs tensorized adapters for LLM adaptation and enhances robustness to data heterogeneity by freezing parts of the tensor factors, thereby substantially reducing trainable parameters while preserving model performance. The work \cite{kim2023client} employs a hypernetwork to generate client-specific adapters, effectively addressing data heterogeneity by enabling on-demand parameter generation tailored to each client. However, these methods do not fundamentally overcome the memory wall, because although they reduce the number of trainable parameters, the activation memory that grows with the context length remains unaffected.

\noindent\textbf{FedLLM with Zeroth-Order Optimization.} To further reduce the memory footprint of FedLLM, many studies adopt BP-free Zeroth-Order Optimization (ZOO) \cite{xu2024fwdllm,qinfederated,panchal2024thinking,malladi2023fine,tan2025harmony}, which replaces backpropagation with forward propagation to reduce the memory consumption \cite{fang2022communication}. 

This work \cite{malladi2023fine} introduces MeZO, a memory-efficient zeroth-order optimizer that adapts classical ZO-SGD to operate in-place, enabling the fine-tuning of extremely large language models with the same memory footprint as inference, supporting both full-parameter and parameter-efficient tuning, and capable of optimizing non-differentiable objectives. This work \cite{tan2025harmony} proposes a divergence-driven zeroth-order optimization method that adapts layer-wise updates to mimic first-order learning patterns, enabling memory-efficient LLM fine-tuning with faster convergence and competitive or superior accuracy compared to both conventional ZO and memory-intensive FO methods. The work \cite{panchal2024thinking} proposes a memory-efficient federated fine-tuning algorithm for large language models by distributing trainable weights across clients and applying forward-mode automatic differentiation (AD) to compute unbiased gradient estimates. The work \cite{qinfederated} applies zeroth-order optimization with a limited set of random seeds, allowing LLM fine-tuning without retaining intermediate activations while also reducing communication overhead. The work \cite{xu2024fwdllm} is another recent BP-free approach for LLMs that relies solely on perturbed inferences rather than full fine-tuning. Nevertheless, the above BP-free methods trade off fine-tuning accuracy for computational efficiency. 

\noindent\textbf{Summary.} The above two types of methods are either constrained by a severe memory footprint, which hinders resource-limited mobile devices from participating in LLM fine-tuning, or affected by cumulative errors in multi-round gradient estimation, which increase convergence time. Consequently, CooperLLM employs a cloud-assisted approach to correct local zeroth-order updates by leveraging gradient knowledge extracted from cloud-based backpropagation, ensuring both computational efficiency on resource-constrained mobile devices and significantly improved convergence stability and accuracy while preserving user data privacy.



\section{Conclusion}
We present CooperLLM, a cloud-assisted federated fine-tuning framework designed to enable efficient and privacy-preserving LLMs adaptation on resource-constrained mobile end devices. By combining the efficiency of ZOO with the accuracy of gradient rectification (ZGR), CooperLLM eliminates the need for costly BP while maintaining convergence stability and task performance. The cloud–end cooperation mechanism allows the cloud to extract and guide gradient subspaces using public auxiliary data, providing directionally informed updates to local client models without exposing private data. Beyond algorithmic efficiency, CooperLLM addresses key system-level bottlenecks in federated LLM fine-tuning. The System-level Pipeline Controller (SPC) overlaps computation and communication through layer-wise pipelining, significantly reducing memory overhead and hiding transmission latency. The Data Transmission Controller (DTC) further minimizes communication delay by applying adaptive quantization and compression to guided perturbations, dynamically balancing transfer and computation latency. Experimental results across multiple Transformer-based models and NLP benchmarks demonstrate that CooperLLM reduces on-device memory footprint by up to 86.37\%, accelerates convergence by up to 8.8×, and improves accuracy by up to 10 percentage points compared to state-of-the-art baselines. These findings confirm CooperLLM as an effective and practical framework for enabling privacy-preserving, resource-efficient, and high-performance fine-tuning of LLMs on mobile end devices.

\bibliographystyle{ACM-Reference-Format}
\bibliography{sample-base}


\end{document}